\documentclass[letterpaper]{article} % DO NOT CHANGE THIS
\usepackage[preprint]{aaai2027}
\usepackage[hyphens]{url}            % DO NOT CHANGE THIS
\usepackage{graphicx}                % DO NOT CHANGE THIS
\usepackage{natbib}                  % DO NOT CHANGE THIS AND DO NOT ADD OPTIONS
\usepackage{caption}                 % DO NOT CHANGE THIS AND DO NOT ADD OPTIONS
\usepackage{booktabs}
\usepackage{amsmath}
\usepackage{amssymb}
\usepackage{algorithm}
\usepackage{algorithmic}

\graphicspath{{figures/}}

\newcommand{\PromptOnlyMarker}{%
  \raisebox{-0.30ex}{\includegraphics[width=0.013\textwidth]{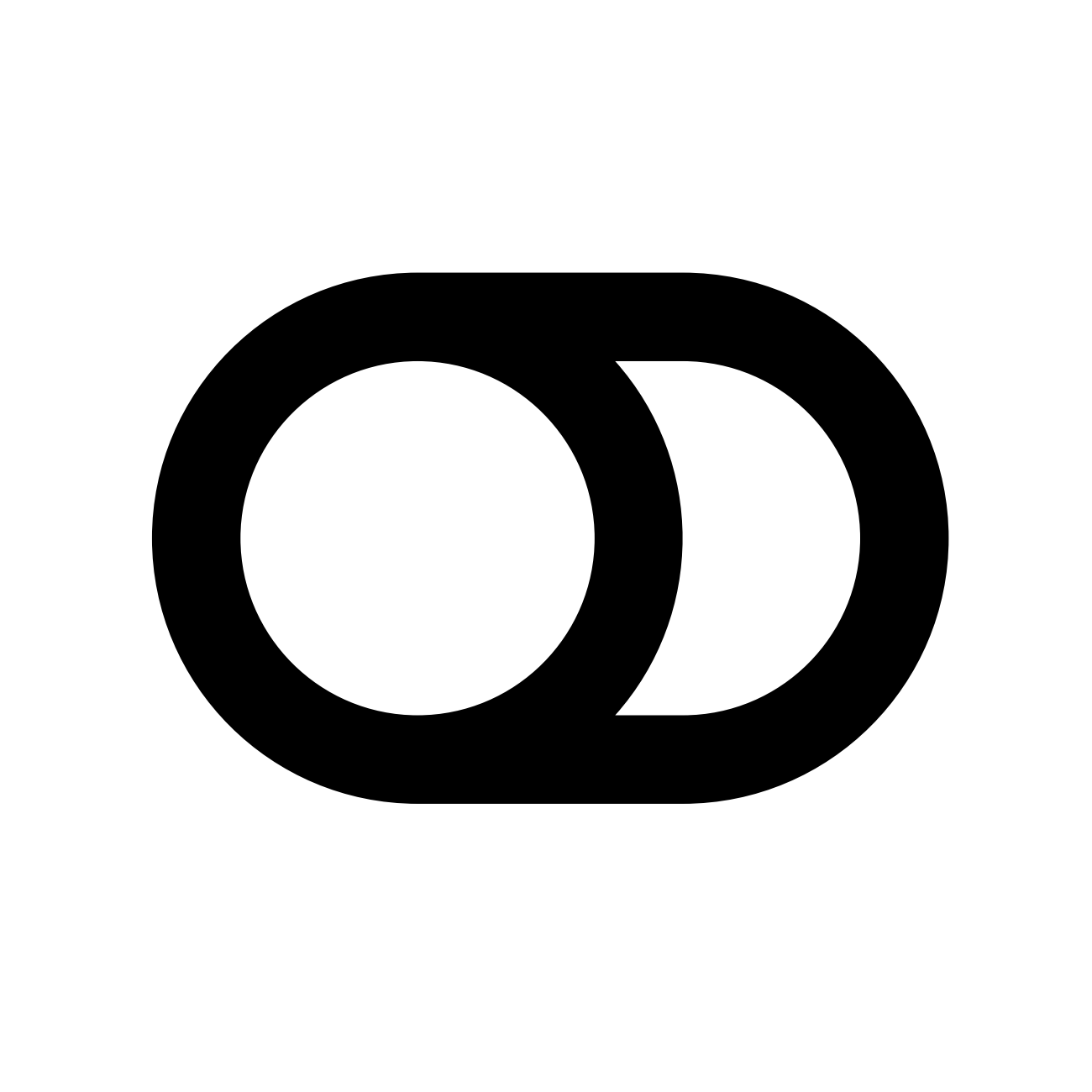}}}
\newcommand{\TrainedMarker}{%
  \raisebox{-0.30ex}{\includegraphics[width=0.013\textwidth]{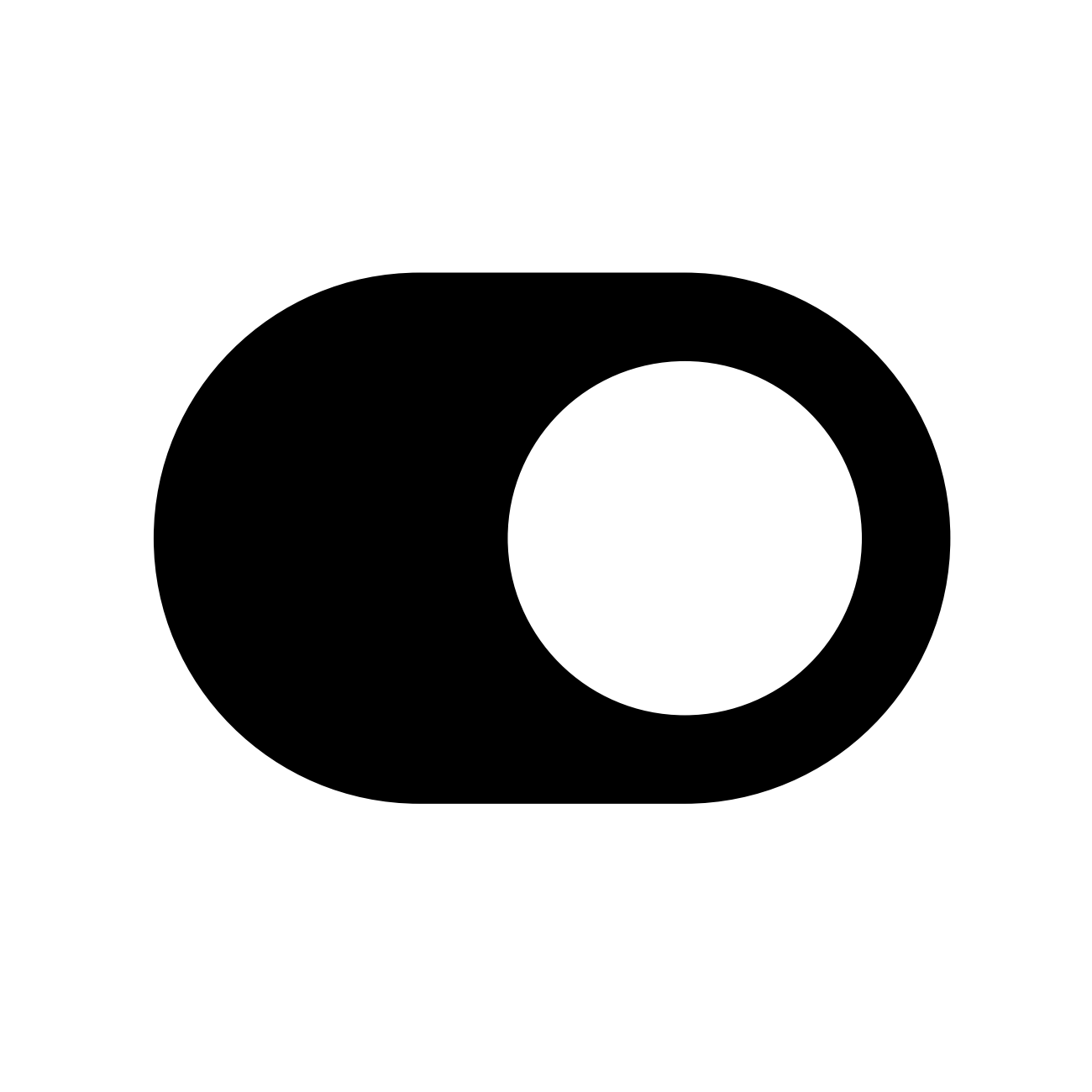}}}
\newcommand{\NoKnowledgeMarker}{%
  \raisebox{-0.30ex}{\includegraphics[width=0.013\textwidth]{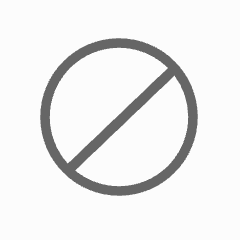}}}
\newcommand{\ChunkKnowledgeMarker}{%
  \raisebox{-0.30ex}{\includegraphics[width=0.013\textwidth]{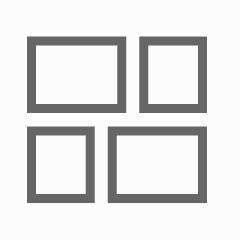}}}
\newcommand{\GraphKnowledgeMarker}{%
  \raisebox{-0.30ex}{\includegraphics[width=0.013\textwidth]{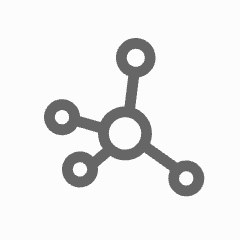}}}

\newcommand{\LLMFreeMarker}{%
  \raisebox{-0.08ex}{\ensuremath{\lozenge}}}
\newcommand{\LLMBasedMarker}{%
  \raisebox{-0.08ex}{\ensuremath{\blacklozenge}}}
\newcommand{\NoConstructionMarker}{%
  \makebox[0.75em][c]{\textcolor{gray}{\textendash}}}
\newcommand{\MethodMarkers}[3]{\mbox{#1#2\kern0.06em#3}\kern0.28em}

\title{Harness-G: A Graph-Structured Harness for Search Agents}
\author{
Yanning Hou\equalcontrib,
Haoyuan Chen\equalcontrib,
Sihang Zhou\corresponding,\\
Xiaoshu Chen,
Xirui Liu,
Duanyang Yuan,
Lingyuan Meng,\\
Siwei Wang,
Quan Liu,
Jian Huang
}
\affiliations{
National University of Defense Technology\\
Changsha, China
}

\begin{document}
\maketitle

\begin{abstract}
Reinforcement learning (RL) search agents commonly model retrieval as
free-form natural-language query generation and optimize multi-turn
interactions using final-answer rewards.
Current studies mainly improve training with denser or more structured
credit signals, but rarely examine whether retrieval is properly
formulated at the policy--environment interface.
We observe pronounced \emph{retrieval aliasing} during Search-R1
training: rollouts for the same question continue to generate distinct
query strings, yet their accumulated evidence sets increasingly
overlap.
We call this phenomenon \emph{retrieval-equivalence collapse}; in this
regime, trajectories approach \emph{utility equivalence} with respect
to retrieval decisions, leaving within-group returns with little
effective retrieval contrast.
To address this problem, we propose \textbf{Harness-G}, a
graph-structured retrieval framework that redesigns this interface.
It reformulates free-form query generation as finite action selection:
the policy selects an evidence sentence or entity, or chooses to answer,
while the environment constructs the menu, tracks retrieval state, and
validates and executes each choice.
This interface reduces linguistic aliasing and makes same-state
alternatives directly comparable.
Building on this interface, we introduce \textbf{Structured Non-myopic
Credit} (\textbf{SNC}), which uses a frozen answer scorer to compare the
selected action with its alternatives and assigns downstream gains to
the earlier actions that enabled them.
Across six QA benchmarks, Harness-G achieves the highest average F1 at
both evaluated model scales, outperforming the strongest baseline,
Graph-R1, by 10.74 points at 1.5B and 3.98 points at 3B.
\end{abstract}

\section{Introduction}

Large language models (LLMs) increasingly rely on external retrieval to
solve knowledge-intensive and multi-hop reasoning tasks
\citep{lewis2020rag,trivedi2023ircot,jiang2023flare}.
Recent reinforcement learning (RL) search agents, such as Search-R1,
train LLMs to generate free-form search queries during reasoning and
optimize complete multi-turn interaction trajectories using final-answer
rewards
\citep{jin2025searchr1,song2025r1searcher,zheng2025deepresearcher}.

Despite substantial progress, their optimization remains brittle,
exhibiting reward crashes, repetitive retrieval, and vanishing
within-group advantages
\citep{igpo2025,wang2025ragen}.
Existing methods primarily address these problems through process
rewards, information gain, or more fine-grained credit signals
\citep{igpo2025,wang2025stepsearch,treegrpo2025,
feng2025gigpo,zhang2025reasonrag}.
Yet they retain the same retrieval interface: the policy
generates a free-form natural-language query, which the environment maps
to retrieved evidence.
We ask whether this interface suits RL optimization.

\begin{figure}[t]
\centering
\includegraphics[width=\columnwidth]{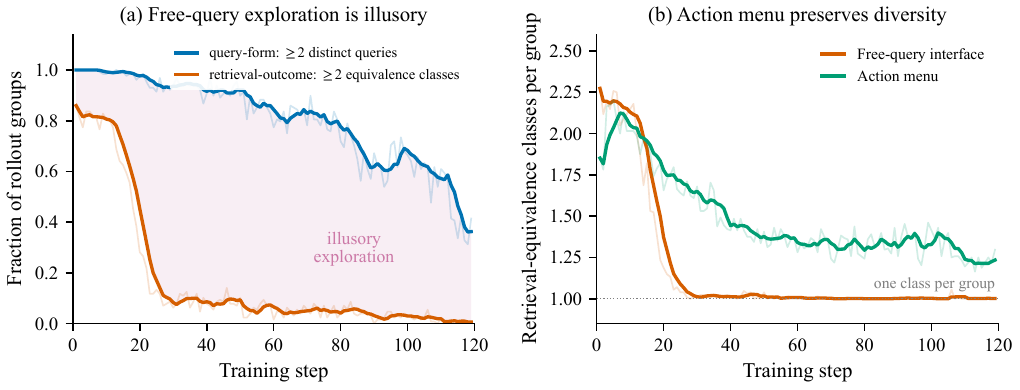}
\caption{\textbf{Retrieval-equivalence collapse.}
(a)~Query-form diversity remains high as retrieval-outcome diversity
collapses, producing illusory exploration.
(b)~Under matched transitions, the action menu preserves more
retrieval-distinct outcomes per rollout group than free querying.}
\label{fig:diagnosis}
\end{figure}

Free-form queries reflect the natural way humans express retrieval
intent through language.
However, LLMs readily generate surface-distinct yet semantically
equivalent queries for the same retrieval intent, so diversity in query
form may not translate into effective diversity in retrieval decisions.
To quantify this mismatch, we measure \emph{query-form diversity}, the
variation among queries generated for the same input, and
\emph{retrieval-outcome diversity}, obtained by clustering trajectories
according to overlap among their accumulated evidence sets; each cluster
defines a \emph{retrieval-equivalence class}.
During Search-R1 training (Figure~\ref{fig:diagnosis}a), query-form
diversity remains substantially above retrieval-outcome diversity, while
the fraction of rollout groups spanning multiple retrieval-equivalence
classes falls from approximately \(86\%\) to below \(10\%\) by training
step \(30\).
The policy thus continues to vary its queries without reaching genuinely
different retrieval outcomes.
We term this \emph{retrieval-equivalence collapse}.

Retrieval-equivalence collapse creates two direct problems for
group-relative optimization.
GRPO infers which actions are preferable from return differences among
trajectories sampled for the same input
\citep{shao2024deepseekmath}; when trajectories gather highly
overlapping evidence, they approach \emph{utility equivalence} with
respect to retrieval decisions.
\textbf{(i) Vanishing advantages.}
If the same evidence leads to the same answer, trajectories receive the
same return and their within-group advantages vanish; nominally distinct
rollouts then provide only a few retrieval-distinct samples.
\textbf{(ii) Confounded retrieval credit.}
If trajectories that retrieve the same evidence produce different
answers, their return differences mainly reflect downstream reasoning or
answer generation rather than retrieval quality; attributing those
differences to retrieval actions cannot reliably identify a better
retrieval choice.
The issue is therefore not only how trajectories are evaluated, but also
whether a rollout group contains genuinely different retrieval
decisions.
Process rewards can provide finer-grained supervision for sampled
trajectories, but they leave the many-to-one mapping from free-form
queries to retrieval outcomes unchanged and therefore cannot
fundamentally restore effective contrast at the retrieval level.

\begin{figure}[t]
\centering
\includegraphics[width=\columnwidth]{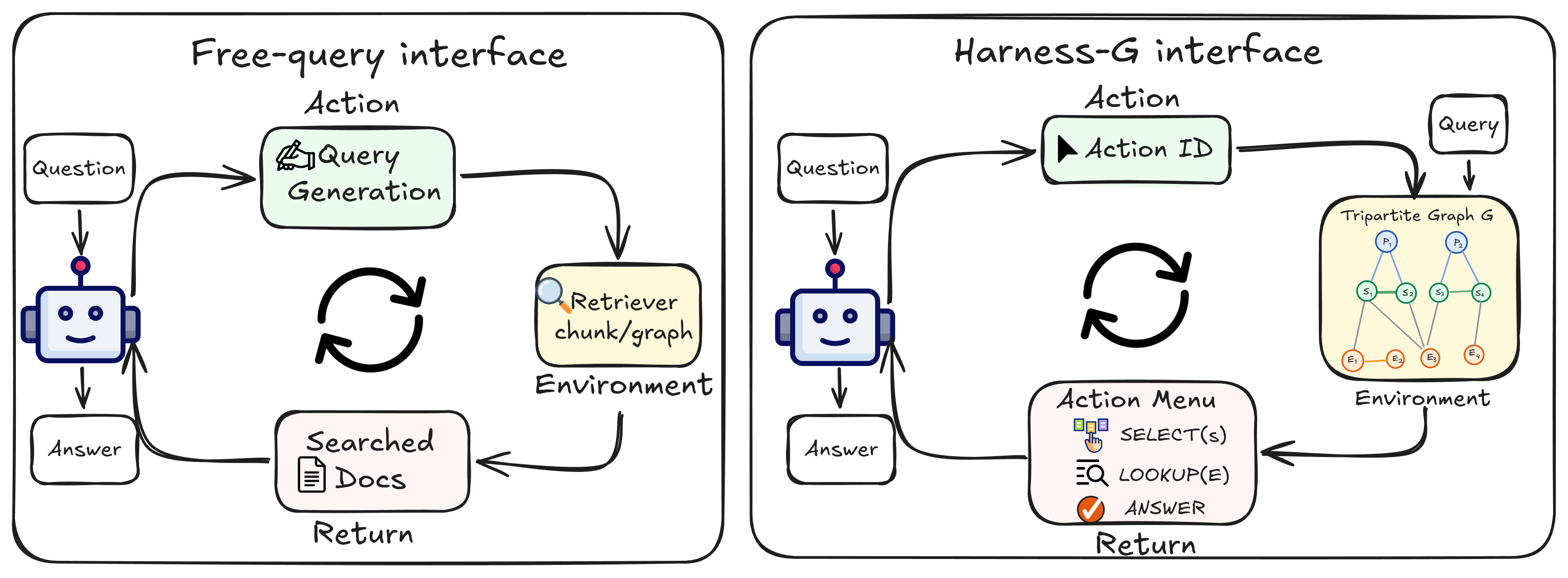}
\caption{\textbf{Retrieval interface redesign.}
Left: free-query interface, where the policy generates a query string and
receives searched documents.
Right: Harness-G interface, where the policy selects a finite action ID from
the menu (\textsc{Select}/\textsc{Lookup}/\textsc{Answer}) over a tripartite
graph, and the environment constructs queries and updates retrieval state.}
\label{fig:intro}
\end{figure}

To this end, we propose \textbf{Harness-G}
(Figure~\ref{fig:intro}), a graph-structured retrieval framework for
search agents.
At its core, Harness-G reformulates free-form query generation as
graph-guided finite action selection: the policy selects a retrieval
target, while the environment constructs queries and updates retrieval
state.
Harness-G organizes the corpus as a paragraph--sentence--entity graph
and exposes a bounded action menu based on the current state.
The policy may commit an evidence sentence (\textsc{Select}), follow an
entity (\textsc{Lookup}), or terminate with an answer
(\textsc{Answer}); the environment validates and deterministically
executes the selected action, including duplicate filtering.
This design makes feasible alternatives at the same decision state
explicit, verifiable, and previewable.
To isolate the effect of how actions are exposed, our matched comparison
holds feasible actions and environment transitions fixed, varying only
whether the policy selects a menu entry directly or reaches the same
target through a generated query (Figure~\ref{fig:diagnosis}b).
Menu selection preserves multiple retrieval-distinct outcomes, whereas
free-query groups rapidly collapse to a single retrieval-equivalence
class.
This result shows that effective exploration depends on retrieval
interface design as well as reward signals.

The resulting action frontier also enables structured step credit.
Building on it, we introduce
\textbf{Structured Non-myopic Credit} (\textbf{SNC}).
First, \emph{frontier-relative credit} uses a frozen answer scorer and
read-only previews to compare the selected action with alternatives at
the same state.
Second, \emph{enablement credit} propagates downstream gains along
trajectory dependencies to the earlier actions that enabled them.
The former provides a local action comparison, while the latter supplies
delayed credit to early bridge actions.

We evaluate Harness-G on six benchmarks spanning multi-hop and
open-domain question answering.
It achieves the highest average F1 at both model scales, outperforming
Graph-R1 by 10.74 points at 1.5B and 3.98 points at 3B; controlled
studies separately validate gains from the action menu and SNC.
Overall, we identify retrieval-equivalence collapse, reformulate
multi-hop retrieval as finite action selection over a graph-structured
environment, and develop SNC for same-state and delayed credit, showing
that both credit assignment and interface design shape search-agent
optimization.

\section{Related Work}

\paragraph{Free-query RL search and credit assignment.}
RL query reformulation predates LLM search agents
\citep{nogueira2017query,buck2018activeqa}; WebGPT and ReAct couple
generation with browser or tool actions
\citep{nakano2021webgpt,yao2023react}.
Search-R1 \citep{jin2025searchr1}, R1-Searcher
\citep{song2025r1searcher}, and DeepResearcher
\citep{zheng2025deepresearcher} train outcome-reward free-query agents;
others add cold starts, live-web research, or simulated retrieval
\citep{chen2025research,li2025webthinker,sun2025zerosearch}.
Training instabilities \citep{wang2025ragen} motivate denser credit:
IGPO \citep{igpo2025} uses per-turn information gain;
StepSearch \citep{wang2025stepsearch} and ReasonRAG
\citep{zhang2025reasonrag} add intermediate supervision; CriticSearch
uses a retrospective critic
\citep{zhang2026criticsearch}; Tree-GRPO
\citep{treegrpo2025} and GiGPO \citep{feng2025gigpo} create step contrast
via tree or anchor-state grouping.
These methods retain free-query targets and hence the many-to-one
string-to-retrieval mapping.
Harness-1 \citep{jiang2026harness1} externalizes search state but leaves
search open-ended at the policy interface; Harness-G instead exposes a
finite, verifiable, and previewable graph menu, enabling direct comparison
among feasible actions and structured non-myopic credit.

\paragraph{Graph retrieval and discrete navigation.}
GraphRAG systems use LLM-built corpus graphs, often for one-shot multi-hop aggregation \citep{edge2024graphrag,guo2024lightrag,gutierrez2024hipporag,gutierrez2025hipporag2,luo2025hypergraphrag}.
Graph-R1 \citep{luo2025graphr1} adds interactive end-to-end RL but retains free-form queries over its graph.
KGQA path agents choose entities or relations on curated schemas \citep{das2018minerva,sun2024tog,kgr12025}, giving finite actions unavailable to open-text free-query agents.
Harness-G brings discrete navigation to open documents: a programmatic, relation-free paragraph--sentence--entity graph exposes a finite, verifiable, and previewable menu with deterministic query construction and state updates.

\section{Method}

\begin{figure*}[t]
\centering
\includegraphics[width=\textwidth]{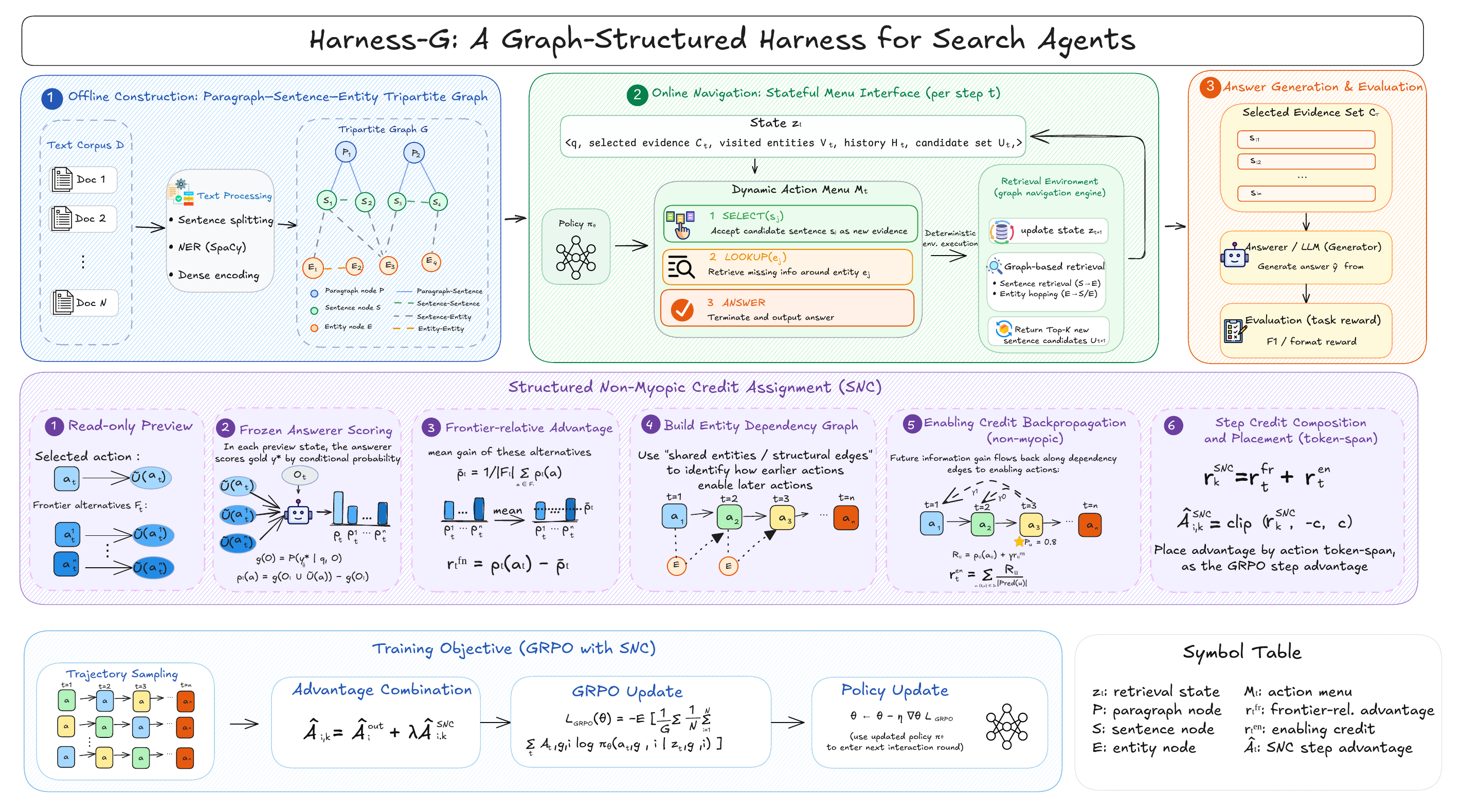}
\caption{Harness-G from graph construction to policy optimization. A paragraph--sentence--entity graph supports the online action menu, where the policy selects finite \textsc{Select}/\textsc{Lookup}/\textsc{Answer} actions and the environment performs deterministic retrieval. SNC previews same-state alternatives, scores them with frozen answerer $g$, and propagates enablement credit along provenance edges; GRPO combines the resulting step advantage with the outcome advantage.}
\label{fig:framework}
\end{figure*}

Harness-G recasts multi-hop retrieval from free-form query generation into \emph{stateful navigation} over a structured evidence space (Figure~\ref{fig:framework}). Offline, it induces a paragraph--sentence--entity graph from the raw corpus. Online, the environment maintains the retrieval state, exposes a finite, deduplicated action menu---adopt an evidence sentence, look up an entity, or answer---and constructs every retrieval query deterministically, leaving the policy one semantic decision per step: which evidence or entity to pursue next. Read-only previews of menu entries then ground Structured Non-myopic Credit (SNC). The essential intervention is not an additional retriever; it is a redefinition of the action space the policy sees.

\subsection{Retrieval as Menu Navigation}

Training data are question--answer pairs $(q,y^*)\sim\mathcal{Q}$ and a corpus $\mathcal{D}=\{p_i\}_{i=1}^{N}$. Under the free-query interface, the action at step $t$ is an arbitrary string $u_t\in\Sigma^*$, whose large retrieval equivalence classes erode the within-group contrast on which group-relative optimizers depend. Harness-G instead defines a finite-horizon decision process of at most $T$ turns. The state collects the question, committed evidence $C_t$, visible candidate sentences $U_t$, visited entities $V_t$, and history $H_t$; the environment exposes a finite menu $M_t$, and the policy selects from it:
\begin{equation}
z_t=\big(q,\;C_t,\;U_t,\;V_t,\;H_t\big),
\qquad
a_t \sim \pi_\theta(\cdot \mid o_t,\,M_t),
\end{equation}
where $a_t\in M_t$ and $o_t$ is the rendered textual observation (committed evidence, visible sentences, executable menu). Actions are discrete operators with explicit semantics and deterministic transitions, not strings.

\subsection{Corpus-Induced Tripartite Graph}

Offline, Harness-G builds a relation-free graph $G$ with paragraph, sentence, and entity nodes. \emph{Paragraph--sentence} edges tie each sentence---the minimal evidence unit---to its document context; \emph{sentence--entity} edges record mentions, so entities bridge cross-sentence and cross-document hops; \emph{sentence--sentence} edges link adjacent sentences, allowing local context expansion without exposing whole paragraphs; \emph{entity--entity} edges connect identical or highly similar entities via surface normalization, title anchors, abbreviation matching, and optional embedding neighbors, and serve only candidate recall. Construction requires sentence splitting, entity recognition and normalization, and dense encoding; it does not invoke a generative LLM for fact/relation extraction or structural organization \citep[cf.][]{luo2025graphr1}. The result is a lightweight, enumerable, traceable interface, built in near-linear time and reused across runs.

\subsection{The Menu Environment}

Each episode opens with an initial retrieval $U_0=\mathrm{InitRetrieve}(q;G)$ that rank-fuses paragraph-local, global-sentence, and entity-mention channels. At every step the policy must emit a menu action id---never a natural-language query---choosing among three action types.

$\textsc{Select}(s)$ with $s\in U_t$ commits a visible sentence, $C_{t+1}=C_t\cup\{s\}$; the final answer conditions only on committed evidence, so \textsc{Select} fixes the evidence state at answer time. $\textsc{Lookup}(e)$ retrieves missing information around entity $e$: the environment---not the model---issues the deterministic query $m_t=\mathrm{concat}(q,\mathrm{text}(C_t))$ against the candidate pool
\begin{equation}
\mathcal{C}(e)=\mathcal{S}(e)\;\cup \bigcup_{e'\in \mathrm{Syn}(e)}\mathcal{S}(e')\;\cup\;\mathcal{N}_{SS}(\mathcal{S}(e)),
\end{equation}
where $\mathcal{S}(e)$ are the sentences mentioning $e$, $\mathrm{Syn}(e)$ its synonym entities, and $\mathcal{N}_{SS}$ sentence-adjacency expansion; the top-$K$ sentences under $\mathrm{sim}(m_t,\cdot)$ become $U_{t+1}$. A \textsc{Lookup} is thus identified by its target entity rather than a generated string, so distinct \textsc{Lookup}s pursue distinct targets by construction. $\textsc{Answer}$ terminates retrieval and generates $\hat y$ from $C_t$. An optional harvest variant $\textsc{Answer\_With}(s_1,\ldots,s_k)$ commits visible sentences and terminates in one step.

The menu is generated as $M_t=\mathrm{Filter}(\mathrm{Propose}(z_t,G),H_t)$: committed sentences and duplicate retrievals are merged away, visited entities never reappear as \textsc{Lookup} targets, and low-value targets (dates, cardinals, nationality adjectives, truncated aliases) are never exposed. Three properties follow by construction. \textbf{Finiteness}: $|M_t|$ is bounded by the visible-sentence and \textsc{Lookup}-candidate caps $K_s$ and $K_e$. \textbf{Verifiability}: every action carries an explicit type and target, so invalid actions are blocked from the feasible set rather than punished after the fact. \textbf{Previewability}: since transitions are deterministic index operations, any candidate $a\in M_t$ can be expanded read-only---$\mathrm{Preview}(z_t,a)$ returns the sentences $\widetilde{U}(a)$ the action would introduce or commit, mutating no real state. Previewability reduces ``how good is this action relative to its genuine same-state alternatives,'' uncomputable over an open string space, to a bounded set of environment operations---the question SNC evaluates.

\subsection{SNC: Structured Non-Myopic Credit}

SNC turns previewability into a low-variance, non-myopic step credit with two complementary terms, both computed entirely by the environment---no external reward model, no extra rollouts \citep[cf.][]{treegrpo2025,feng2025gigpo}.

\paragraph{Frontier-relative advantage.}
A frozen answerer scores how well the observed evidence supports the gold answer, and previews give each information-acquisition action a marginal gain:
\begin{equation}
g(O)=P_{\bar\theta}(y^*\mid q,O),
\qquad
p_t(a)=g\big(O_t\cup \widetilde{U}(a)\big)-g(O_t),
\label{eq:gain}
\end{equation}
where $g$ is the length-normalized teacher-forced gold-answer probability under stop-gradient parameters $\bar\theta$, maximized over gold aliases (appendix), and $O_t\supseteq C_t$ is the scorer-only cumulative context containing every sentence surfaced before $a_t$ in stable first-seen order. In contrast, $C_t$ remains the committed evidence used for lookup queries and final answering. Thus, re-surfacing observed text gains exactly zero. For an information-acquisition action $a_t$, SNC also previews a bounded counterfactual frontier $\mathcal{F}_t\subseteq M_t\setminus\{a_t\}$ of information-acquiring alternatives (type-stratified, deduplicated by resulting scorer context, capped at $K_f$). We denote the mean gain of these alternatives by
\begin{equation}
\bar p_t=\frac{1}{|\mathcal{F}_t|}\sum_{a\in\mathcal{F}_t}p_t(a),
\label{eq:frontier-baseline}
\end{equation}
Subtracting this in-menu baseline gives the frontier-relative advantage:
\begin{equation}
r_t^{\mathrm{fr}}=p_t(a_t)-\bar p_t.
\label{eq:frontier}
\end{equation}
For a non-information action (commit or terminal), or when $\mathcal{F}_t=\varnothing$, we set $r_t^{\mathrm{fr}}=0$.
When several menu entries would add equivalent evidence, their gains are close and $r_t^{\mathrm{fr}}\approx 0$: the policy is not reinforced for a superficial choice; positive credit requires outperforming feasible alternatives at the same state, a comparison unavailable in an open query space.

\paragraph{Structured enablement credit.}
Early hops often pay off only later---they surface the bridge entity whose attribute a subsequent \textsc{Lookup} retrieves---so per-step marginal gains systematically undervalue them. SNC records a \emph{provenance} edge $(t,u)\in\mathcal{D}_\tau$ ($t<u$, hence acyclic) whenever step $u$ consumes a sentence or entity that step $t$ produced, and propagates credit over this graph in reverse topological order:
\begin{equation}
R_u=p_u(a_u)+\gamma\,r_u^{\mathrm{en}},
\qquad
r_t^{\mathrm{en}}=\sum_{u:\,(t,u)\in\mathcal{D}_\tau} \frac{R_u}{|\mathrm{Pred}(u)|},
\label{eq:enable}
\end{equation}
with $\mathrm{Pred}(u)$ the direct producers of step $u$ and $\gamma$ a propagation discount ($\gamma{=}1$ by default). Downstream gains thus flow back through entire enabling chains, teaching the non-myopic pattern ``first find the bridge entity, then look up its missing attribute.'' We apply a small dead-zone to $p_t$ against scoring noise. The total step credit is then $r_t^{\mathrm{SNC}}=r_t^{\mathrm{fr}}+r_t^{\mathrm{en}}$.

\begin{table*}[t]
\centering
\small
\setlength{\tabcolsep}{2.0pt}
\renewcommand{\arraystretch}{1.04}
\footnotesize
\begin{tabular}{@{}l *{16}{c}@{}}
\toprule
& \multicolumn{12}{c}{Per-dataset} & \multicolumn{4}{c}{Average} \\
\cmidrule(lr){2-13}\cmidrule(l){14-17}
& \multicolumn{2}{c}{2Wiki} & \multicolumn{2}{c}{HotpotQA} & \multicolumn{2}{c}{MuSiQue} & \multicolumn{2}{c}{NQ} & \multicolumn{2}{c}{PopQA} & \multicolumn{2}{c}{TriviaQA} & \multicolumn{4}{c}{Avg.} \\
\cmidrule(lr){2-3}\cmidrule(lr){4-5}\cmidrule(lr){6-7}\cmidrule(lr){8-9}\cmidrule(lr){10-11}\cmidrule(lr){12-13}\cmidrule(l){14-17}
Method
& F1 & G-E & F1 & G-E & F1 & G-E & F1 & G-E & F1 & G-E & F1 & G-E
& EM & F1 & R-S & G-E \\
\midrule
\multicolumn{17}{@{}l}{\textit{GPT-4o-mini}} \\
\addlinespace[1pt]
\MethodMarkers{\PromptOnlyMarker}{\NoKnowledgeMarker}{\NoConstructionMarker}Naive generation
& 17.03 & 74.86 & 31.79 & 78.48 & 11.45 & 76.61 & 21.59 & 84.64 & 25.95 & 72.75 & 47.73 & 83.33
& 11.36 & 25.92 & -- & 78.45 \\
\MethodMarkers{\PromptOnlyMarker}{\ChunkKnowledgeMarker}{\LLMFreeMarker}Standard RAG
& 22.31 & 73.02 & 46.70 & 81.88 & 17.31 & 74.93 & 26.85 & 84.55 & 30.58 & 69.42 & 48.55 & 84.63
& 18.10 & 32.05 & 52.68 & 78.07 \\
\MethodMarkers{\PromptOnlyMarker}{\GraphKnowledgeMarker}{\LLMBasedMarker}GraphRAG
& 16.02 & 72.81 & 31.67 & 77.37 & 15.14 & 74.43 & 20.31 & 82.36 & 20.92 & 65.88 & 45.13 & 82.76
& 12.50 & 24.87 & 32.48 & 75.94 \\
\MethodMarkers{\PromptOnlyMarker}{\GraphKnowledgeMarker}{\LLMBasedMarker}LightRAG
& 16.59 & 71.94 & 30.70 & 73.42 & 14.39 & 73.75 & 19.09 & 80.20 & 20.47 & 67.76 & 40.18 & 81.60
& 9.77 & 23.57 & 47.42 & 74.78 \\
\MethodMarkers{\PromptOnlyMarker}{\GraphKnowledgeMarker}{\LLMBasedMarker}PathRAG
& 12.42 & 67.19 & 23.12 & 71.81 & 11.49 & 69.94 & 20.01 & 81.99 & 15.65 & 60.58 & 37.44 & 80.94
& 7.03 & 20.02 & 46.71 & 72.08 \\
\MethodMarkers{\PromptOnlyMarker}{\GraphKnowledgeMarker}{\LLMBasedMarker}HippoRAG2
& 16.27 & 68.78 & 31.78 & 76.43 & 12.37 & 73.05 & 24.56 & 84.65 & 21.10 & 63.31 & 46.86 & 83.55
& 13.80 & 25.49 & 36.41 & 74.96 \\
\MethodMarkers{\PromptOnlyMarker}{\GraphKnowledgeMarker}{\LLMBasedMarker}HyperGraphRAG
& 21.14 & 76.76 & 37.46 & 80.50 & 20.40 & 79.29 & 22.95 & 81.22 & 29.48 & 70.55 & 44.95 & 85.20
& 13.15 & 29.40 & 61.82 & 78.92 \\
\midrule
\multicolumn{17}{@{}l}{\textit{Qwen2.5-1.5B-Instruct}} \\
\addlinespace[1pt]
\MethodMarkers{\PromptOnlyMarker}{\NoKnowledgeMarker}{\NoConstructionMarker}Naive generation
& 7.78 & 49.13 & 4.27 & 45.77 & 2.35 & 46.63 & 6.03 & 46.74 & 10.06 & 42.67 & 8.10 & 52.92
& 1.17 & 6.43 & -- & 47.31 \\
\MethodMarkers{\PromptOnlyMarker}{\ChunkKnowledgeMarker}{\LLMFreeMarker}Standard RAG
& 11.46 & 55.38 & 9.93 & 52.91 & 3.18 & 39.46 & 11.39 & 59.73 & 13.08 & 50.29 & 17.43 & 60.52
& 5.73 & 11.08 & 52.84 & 53.05 \\
\MethodMarkers{\TrainedMarker}{\NoKnowledgeMarker}{\NoConstructionMarker}SFT
& 13.26 & 34.72 & 13.61 & 38.93 & 5.14 & 28.50 & 11.56 & 46.61 & 15.61 & 31.35 & 26.18 & 46.66
& 9.83 & 14.23 & -- & 37.80 \\
\MethodMarkers{\TrainedMarker}{\NoKnowledgeMarker}{\NoConstructionMarker}R1 (GRPO)
& 26.28 & 47.48 & 20.07 & 44.43 & 4.84 & 39.12 & 16.75 & 45.95 & 21.36 & 44.50 & 34.78 & 48.59
& 14.19 & 20.68 & -- & 45.01 \\
\MethodMarkers{\TrainedMarker}{\ChunkKnowledgeMarker}{\LLMFreeMarker}Search-R1
& 28.43 & 60.61 & 39.99 & 64.16 & 4.69 & 39.32 & 20.26 & 59.93 & 39.63 & 58.19 & 44.16 & 63.01
& 23.18 & 29.53 & 50.45 & 57.54 \\
\MethodMarkers{\TrainedMarker}{\ChunkKnowledgeMarker}{\LLMFreeMarker}IGPO
& 32.48 & 63.45 & 44.15 & 67.28 & 9.02 & 43.15 & 24.08 & 62.48 & 43.52 & 61.35 & 48.21 & 66.12
& 26.68 & 33.58 & 53.65 & 60.64 \\
\MethodMarkers{\TrainedMarker}{\ChunkKnowledgeMarker}{\LLMFreeMarker}R1-Searcher
& 28.01 & 58.81 & \underline{41.50} & 61.54 & 6.26 & 38.31 & \textbf{36.86} & \underline{60.79} & 38.37 & 56.02 & 42.57 & 61.24
& 23.70 & 32.26 & 50.68 & 56.12 \\
\MethodMarkers{\TrainedMarker}{\GraphKnowledgeMarker}{\LLMBasedMarker}Graph-R1
& \underline{35.13} & \underline{65.73} & 40.62 & \underline{65.30} & \underline{28.28} & \underline{58.82} & 35.62 & 59.13 & \underline{43.55} & \underline{66.46} & \underline{57.36} & \underline{70.83}
& \underline{31.90} & \underline{40.09} & \underline{59.35} & \underline{64.38} \\
\MethodMarkers{\TrainedMarker}{\GraphKnowledgeMarker}{\LLMFreeMarker}\textbf{Harness-G}
& \textbf{64.59} & \textbf{78.14} & \textbf{56.33} & \textbf{77.19} & \textbf{38.90} & \textbf{65.30} & \underline{35.81} & \textbf{68.73} & \textbf{48.54} & \textbf{72.29} & \textbf{60.80} & \textbf{76.77}
& \textbf{43.10} & \textbf{50.83} & \textbf{64.53} & \textbf{73.07} \\
\midrule
\multicolumn{17}{@{}l}{\textit{Qwen2.5-3B-Instruct}} \\
\addlinespace[1pt]
\MethodMarkers{\PromptOnlyMarker}{\NoKnowledgeMarker}{\NoConstructionMarker}Naive generation
& 7.59 & 55.00 & 11.16 & 53.75 & 3.67 & 54.00 & 8.90 & 57.18 & 10.89 & 49.08 & 10.89 & 48.16
& 3.26 & 8.85 & -- & 52.86 \\
\MethodMarkers{\PromptOnlyMarker}{\ChunkKnowledgeMarker}{\LLMFreeMarker}Standard RAG
& 12.52 & 60.01 & 15.41 & 62.51 & 2.92 & 50.40 & 10.69 & 65.13 & 14.70 & 57.25 & 21.92 & 68.43
& 3.39 & 13.03 & 52.69 & 60.62 \\
\MethodMarkers{\TrainedMarker}{\NoKnowledgeMarker}{\NoConstructionMarker}SFT
& 12.40 & 52.31 & 16.48 & 51.35 & 5.04 & 51.31 & 11.23 & 58.20 & 16.95 & 46.42 & 33.02 & 59.98
& 9.64 & 15.85 & -- & 53.26 \\
\MethodMarkers{\TrainedMarker}{\NoKnowledgeMarker}{\NoConstructionMarker}R1 (GRPO)
& 28.45 & 56.92 & 25.33 & 55.38 & 8.07 & 47.53 & 21.51 & 55.11 & 27.11 & 48.65 & 47.91 & 60.74
& 19.66 & 26.40 & -- & 54.06 \\
\MethodMarkers{\TrainedMarker}{\ChunkKnowledgeMarker}{\LLMFreeMarker}Search-R1
& 38.04 & 54.39 & 43.84 & 69.32 & 7.65 & 46.43 & 37.96 & 52.90 & 38.67 & 63.74 & 47.99 & 60.37
& 28.65 & 35.69 & 49.99 & 57.86 \\
\MethodMarkers{\TrainedMarker}{\ChunkKnowledgeMarker}{\LLMFreeMarker}IGPO
& 42.81 & 58.12 & 48.52 & 71.85 & 12.18 & 50.21 & 41.52 & 56.48 & 42.15 & 66.32 & 51.68 & 63.95
& 32.40 & 39.81 & 53.82 & 61.16 \\
\MethodMarkers{\TrainedMarker}{\ChunkKnowledgeMarker}{\LLMFreeMarker}R1-Searcher
& 23.50 & 55.86 & 42.44 & 64.60 & 12.81 & 50.07 & 36.53 & 63.33 & 40.18 & 66.23 & 54.00 & 60.52
& 27.08 & 34.91 & 49.98 & 60.10 \\
\MethodMarkers{\TrainedMarker}{\GraphKnowledgeMarker}{\LLMBasedMarker}Graph-R1
& \underline{57.56} & \underline{76.45} & \underline{56.75} & \underline{77.46} & \underline{40.51} & \underline{67.84} & \textbf{44.75} & \underline{69.92} & \underline{45.65} & \underline{71.27} & \underline{62.31} & \underline{75.01}
& \underline{42.45} & \underline{51.26} & \underline{60.19} & \underline{72.99} \\
\MethodMarkers{\TrainedMarker}{\GraphKnowledgeMarker}{\LLMFreeMarker}\textbf{Harness-G}
& \textbf{65.53} & \textbf{79.74} & \textbf{65.87} & \textbf{81.52} & \textbf{46.46} & \textbf{71.55} & \underline{42.91} & \textbf{72.85} & \textbf{47.27} & \textbf{77.67} & \textbf{63.40} & \textbf{81.34}
& \textbf{47.14} & \textbf{55.24} & \textbf{64.55} & \textbf{77.44} \\
\bottomrule
\end{tabular}
\small
\caption{Main results with best in \textbf{bold} and second \underline{underlined}.
Markers denote three method properties: optimization regime (\PromptOnlyMarker{} prompt-only vs.\ \TrainedMarker{} trained with RL/SFT), knowledge interface (\NoKnowledgeMarker{} parametric only, \ChunkKnowledgeMarker{} retrieved text chunks, \GraphKnowledgeMarker{} graph-structured knowledge), and construction regime (\LLMFreeMarker{} programmatic/LLM-free, \LLMBasedMarker{} LLM-based extraction or organization, \NoConstructionMarker{} not applicable).}
\label{tab:main}
\end{table*}

\paragraph{Integration with GRPO.}
Each rollout receives $R_i^{\mathrm{out}}=\max_{y\in\mathcal{Y}_i}\mathrm{F1}(\operatorname{norm}(\hat y_i),\operatorname{norm}(y))\in[0,1]$, the per-trajectory token-overlap F1 over accepted answer aliases (Appendix~\ref{app:metrics}); GRPO group-normalizes these rewards into $\hat A_i^{\mathrm{out}}$ \citep{shao2024deepseekmath}. SNC neither replaces nor mixes into the outcome reward: $r_t^{\mathrm{SNC}}$ is spread uniformly over the response tokens of step $t$, normalized by a single batch-global scale (the standard deviation of nonzero SNC token credits, floored at $s_{\min}$; no mean subtraction, so signs are preserved), and clipped to $[-c,c]$, giving $\hat A_{i,k}^{\mathrm{SNC}}$. The token-level advantage is the additive two-stream combination
\begin{equation}
\hat A_{i,k}=\hat A_i^{\mathrm{out}}+\lambda\,\hat A_{i,k}^{\mathrm{SNC}},
\label{eq:combine}
\end{equation}
optimized with the standard clipped surrogate and KL regularization; environment-injected observation tokens are loss-masked and receive no credit. Outcome correctness thus stays primary, while local credit for evidence and entity choices lands on their action tokens. Algorithm~\ref{alg:training} details training.

\begin{figure*}[t]
\centering
\begin{minipage}[t]{0.48\textwidth}
\centering
\includegraphics[width=\linewidth]{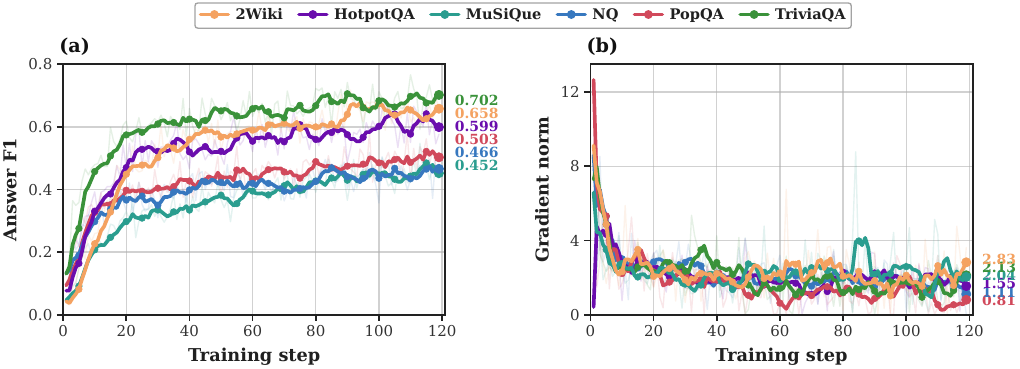}
\caption{Training dynamics of Harness-G (Qwen2.5-3B): (a)~training-batch F1; (b)~gradient norm.}
\label{fig:menu-dynamics}
\end{minipage}\hfill
\begin{minipage}[t]{0.25\textwidth}
\centering
\includegraphics[width=\linewidth]{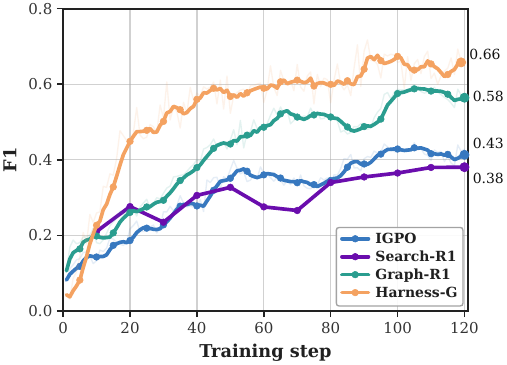}
\caption{F1 on 2Wiki during training (Qwen2.5-3B).}
\label{fig:method-f1}
\end{minipage}\hfill
\begin{minipage}[t]{0.25\textwidth}
\centering
\includegraphics[width=\linewidth]{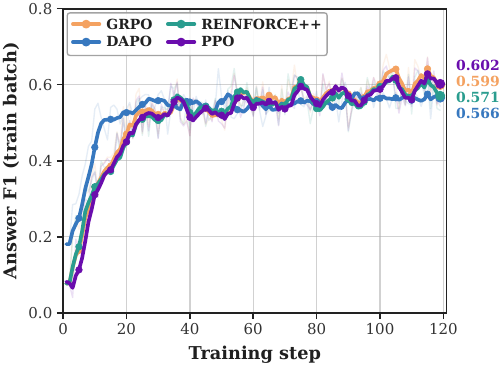}
\caption{Training-batch F1 under different RL algorithms.}
\label{fig:rl-estimators}
\end{minipage}
\end{figure*}

\section{Experiments}

This section presents the experimental setup, main results, and analyses. We answer the following research questions (RQs): \textbf{RQ1}: Does Harness-G outperform existing methods? \textbf{RQ2}: Do the action menu and SNC contribute to performance? \textbf{RQ3}: How stable and robust is Harness-G across datasets, model families, and RL algorithms? \textbf{RQ4}: How interaction-efficient is Harness-G? \textbf{RQ5}: How well does Harness-G generalize under O.O.D.\ settings?

\subsection{Experimental Setup}

\paragraph{Datasets.}
Following Graph-R1 \citep{luo2025graphr1}, we conduct experiments on six standard RAG benchmarks: three multi-hop datasets---2WikiMultiHopQA \citep{ho20202wiki}, HotpotQA \citep{yang2018hotpotqa}, and MuSiQue \citep{trivedi2022musique}---and three single-hop datasets---Natural Questions (NQ) \citep{kwiatkowski2019nq}, PopQA \citep{mallen2023popqa}, and TriviaQA \citep{joshi2017triviaqa}.
Dataset and split details are provided in Appendix~\ref{app:data}.

\paragraph{Baselines.}
We compare Harness-G with Naive Generation, Standard RAG \citep{lewis2020rag}, SFT, R1 \citep{deepseekr1}, Search-R1 \citep{jin2025searchr1}, IGPO \citep{igpo2025}, R1-Searcher \citep{song2025r1searcher}, and Graph-R1 \citep{luo2025graphr1} under Qwen2.5-1.5B/3B \citep{qwen25}, and with GPT-4o-mini-based GraphRAG \citep{edge2024graphrag}, LightRAG \citep{guo2024lightrag}, PathRAG \citep{chen2025pathrag}, HippoRAG2 \citep{gutierrez2025hipporag2}, and HyperGraphRAG \citep{luo2025hypergraphrag}.
Graph-based baselines use GPT-4o-mini uniformly for knowledge construction.
Table~\ref{tab:main} summarizes optimization, knowledge interface, and construction.

% Declare Table~2 after the page-5 right column has started so it lands in
% the next available column: the left column on page~6.
\begin{table}[t]
\centering
\small
\setlength{\tabcolsep}{3.5pt}
\renewcommand{\arraystretch}{1.12}
\begin{tabular}{@{}llccc@{}}
\toprule
Interface & Step credit & 2Wiki & Hotpot & MuSiQue \\
\midrule
Free-query & None
& 38.04/32.03
& 43.84/36.72
& 7.65/4.69 \\
Menu & None
& 62.45/54.32
& 61.32/54.21
& 43.58/32.78 \\
Free-query & IGPO
& 42.81/35.94
& 48.52/40.62
& 12.18/7.81 \\
Menu & IGPO
& 63.58/55.47
& 62.74/55.47
& 44.71/33.59 \\
\midrule
Menu (Ours) & SNC
& \textbf{65.53/59.38}
& \textbf{65.87/59.38}
& \textbf{46.46/35.16} \\
\bottomrule
\end{tabular}
\caption{Multi-hop QA by interface and credit (F1/EM).}
\label{tab:interface}
\end{table}

\paragraph{Evaluation Metrics.}
We evaluate all methods with Exact Match (EM), F1, retrieval similarity (R-S), and generation evaluation (G-E), following Graph-R1 \citep{luo2025graphr1}.
Dataset results show F1 and G-E; averages cover all metrics.

\paragraph{Implementation Details.}
Harness-G uses Qwen2.5-1.5B/3B-Instruct \citep{qwen25} and GRPO \citep{shao2024deepseekmath} with a group size of 8, batch size of 128, 120 training steps, and at most 6 interaction turns.
The outcome reward is token-level answer F1, with environment-injected retrieval tokens loss-masked. Appendix~\ref{app:impl} lists all hyperparameters.

\subsection{Main Results (RQ1)}

As shown in Table~\ref{tab:main}, Harness-G achieves the highest average F1 at both model scales. We make two key observations.

\paragraph{Structured Navigation Particularly Benefits Multi-Hop QA.}
With Qwen2.5-3B, Harness-G achieves 55.24 average F1, outperforming Graph-R1 by 3.98 points.
The improvements are most pronounced on 2Wiki, HotpotQA, and MuSiQue, where F1 increases by 7.97, 9.12, and 5.95 points, respectively (7.68 on average).
Harness-G also obtains higher G-E on all six datasets and improves average R-S from 60.19 to 64.55, although it trails Graph-R1 by 1.84 F1 on NQ.

\paragraph{Smaller Models Gain More from Harness-G.}
At 1.5B, Harness-G raises average F1 from 40.09 to 50.83, a 10.74-point improvement over Graph-R1, and outperforms it on every dataset.
The only higher per-dataset score in this block is R1-Searcher on NQ (36.86 vs.\ 35.81).
These results indicate that restricting retrieval to executable, structured actions is particularly helpful at limited model capacity.

% Declare early so the figure* pair lands on the next page (p.6) rather than p.7.

\subsection{Ablation and Comparative Analysis (RQ2)}

\begin{figure}[t]
\centering
% Scale uniformly to column width; never set both width and height.
\includegraphics[width=\columnwidth,keepaspectratio]{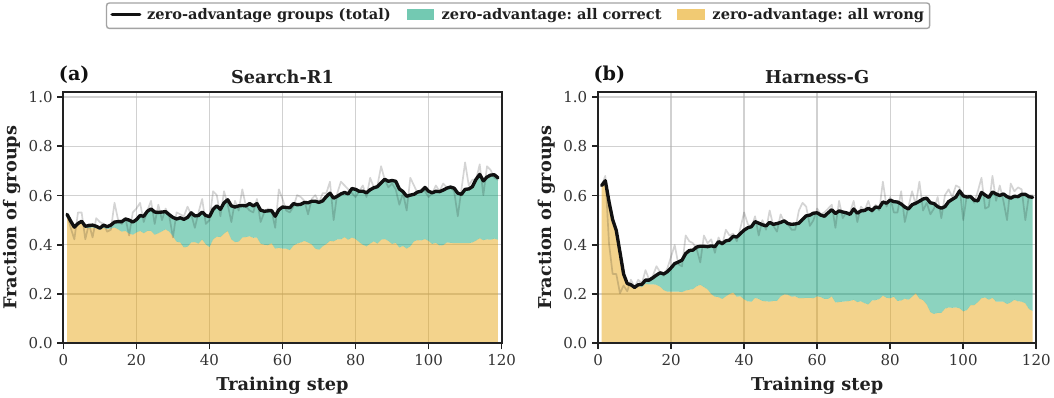}
\caption{Zero-advantage groups on 2Wiki (Qwen2.5-3B): (a)~Search-R1; (b)~Harness-G. Stacked regions separate all-wrong from all-correct groups; the curve shows their total.}
\label{fig:zadv}
\end{figure}

Tables~\ref{tab:interface}--\ref{tab:snc} and Figures~\ref{fig:diagnosis} and~\ref{fig:zadv} ablate the interface and step credit under a fixed graph, outcome reward, training budget, and GRPO configuration. Appendix~\ref{app:protocols} details their effects on retrieval diversity and zero-advantage groups.

\paragraph{Menu vs.\ Free-Query Interface.}
The action menu improves F1 by more than $17$ points over free-query under both credit regimes, and by more than $35$ points on MuSiQue under outcome-only training.
IGPO \citep{igpo2025} densifies free-query credit but leaves a large residual gap to the menu; Menu+SNC is best on all three multi-hop datasets.
Figure~\ref{fig:diagnosis}b shows free-query rollouts collapsing into few retrieval-equivalence classes under the interface-only change, while menu rollouts remain retrieval-diverse.

\paragraph{Zero-Advantage Groups under GRPO.}
Unlike Search-R1, Harness-G preserves retrieval diversity rather than collapsing early (Figures~\ref{fig:zadv} and~\ref{fig:diagnosis}b).
Its zero-advantage rate bottoms at $23\%$ near step~10; the subsequent rebound is driven by all-correct rather than all-wrong groups, indicating successful convergence after diverse exploration.

\paragraph{SNC Ablation.}
As shown in Table~\ref{tab:snc}, full SNC is best on every multi-hop dataset.
Removing both terms ($\lambda{=}0$) drops F1 by $3.08$, $4.55$, and $2.88$ on 2Wiki, HotpotQA, and MuSiQue; either removal also hurts.

\begin{table}[t]
\centering
\small
\setlength{\tabcolsep}{2.4pt}
\renewcommand{\arraystretch}{1.10}
\resizebox{\columnwidth}{!}{%
\begin{tabular}{@{}l*{8}{c}@{}}
\toprule
& \multicolumn{2}{c}{Credit?}
& \multicolumn{2}{c}{2Wiki}
& \multicolumn{2}{c}{Hotpot}
& \multicolumn{2}{c}{MuSiQue} \\
\cmidrule(lr){2-3}\cmidrule(lr){4-5}\cmidrule(lr){6-7}\cmidrule(lr){8-9}
Variant
& $r^{\mathrm{fr}}$ & $r^{\mathrm{en}}$
& F1 & EM
& F1 & EM
& F1 & EM \\
\midrule
Full (Ours)
& $\checkmark$ & $\checkmark$
& \textbf{65.53} & \textbf{59.38}
& \textbf{65.87} & \textbf{59.38}
& \textbf{46.46} & \textbf{35.16} \\
\midrule
w/o Enable.
& $\checkmark$ &
& 63.42 & 55.33
& 63.65 & 53.50
& 45.32 & 34.50 \\
w/o Frontier
& & $\checkmark$
& 64.22 & 57.62
& 63.57 & 57.36
& 46.02 & 34.50 \\
w/o SNC ($\lambda{=}0$)
& &
& 62.45 & 54.32
& 61.32 & 54.21
& 43.58 & 32.78 \\
\bottomrule
\end{tabular}%
}
\caption{SNC leave-one-out ablation on the action menu.}
\label{tab:snc}
\end{table}

% Fig.~5 + Table~6 side-by-side; declared after Tables~2--3 so numbering is 5/6 and the pair tops p.7.
\begin{figure*}[t]
\centering
\begin{minipage}[t]{0.54\textwidth}
\vspace{0pt}
\centering
\includegraphics[width=\linewidth]{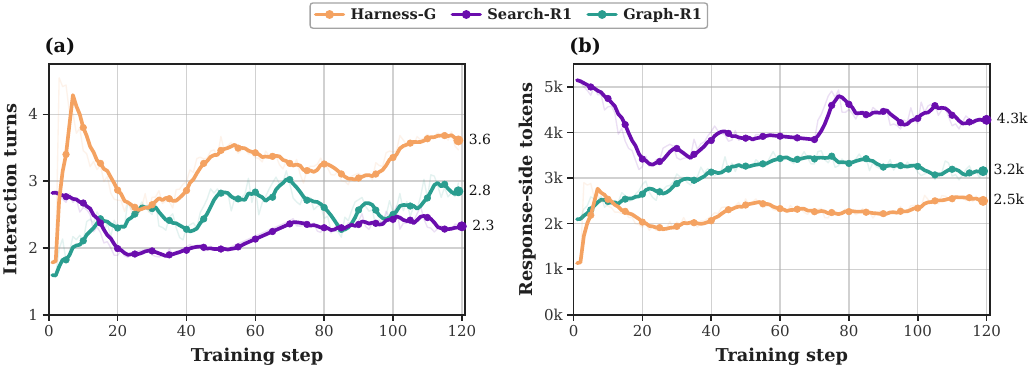}
\caption{Interaction turns (a) and response length (b) during Qwen2.5-3B training on 2Wiki. By late training, Harness-G takes more turns than Search-R1 and Graph-R1 but gives shorter responses.}
\label{fig:turns-length}
\end{minipage}\hfill
\begin{minipage}[t]{0.44\textwidth}
\vspace{0pt}
\centering
\scriptsize
\setlength{\tabcolsep}{1.6pt}
\renewcommand{\arraystretch}{0.90}
\captionsetup{skip=4pt}
% Force Table~6 (declared early for page-7 placement; Tables~4--5 follow in source).
\setcounter{table}{5}%
\resizebox{0.90\linewidth}{!}{%
\begin{tabular}{@{}l cccc cc c@{}}
\toprule
& \multicolumn{4}{c}{Construction} & \multicolumn{2}{c}{Query} & \\
\cmidrule(lr){2-5}\cmidrule(lr){6-7}
Method & TP1KT & CP1MT & \#N & \#E & TPQ & CP1KQ & F1 \\
\midrule
Naive gen. & 0\,s & \$0 & -- & -- & 3.7\,s & \$0.16 & 17.0 \\
Std.\ RAG & 0\,s & \$0 & -- & -- & 4.1\,s & \$1.35 & 22.3 \\
GraphRAG & 8.04\,s & \$3.35 & 7.8k & 4.9k & 7.4\,s & \$3.97 & 16.0 \\
LightRAG & 6.84\,s & \$4.07 & 59k & 25k & 12.2\,s & \$8.11 & 16.6 \\
PathRAG & 6.84\,s & \$4.07 & 59k & 25k & 15.8\,s & \$8.28 & 12.4 \\
HippoRAG2 & 3.25\,s & \$1.26 & 12k & 41k & 8.8\,s & \$7.68 & 16.3 \\
HyperGraphRAG & 6.76\,s & \$4.14 & 174k & 114k & 9.6\,s & \$8.76 & 21.1 \\
Graph-R1 (7B) & 5.69\,s & \$2.81 & 120k & 98k & 7.0\,s & \$0 & 65.0 \\
\midrule
\textbf{Harness-G (3B)} & 0.12\,s & \textbf{\$0} & 242k & 763k & 5.1\,s & \textbf{\$0} & \textbf{65.53} \\
\bottomrule
\end{tabular}%
}
\captionof{table}{Time \& Cost Comparisons on 2Wiki.}
\label{tab:build}
\setcounter{table}{3}%
\end{minipage}
\end{figure*}

\subsection{Training Stability and Robustness (RQ3)}

As shown in Figures~\ref{fig:menu-dynamics}--\ref{fig:rl-estimators} and Table~\ref{tab:robust-family}, Harness-G trains stably and transfers across backbones and RL algorithms.

\paragraph{Stable Optimization across Datasets.}
Across all six 3B runs, training-batch F1 rises without persistent late-stage collapse, with bounded gradient norms (Figure~\ref{fig:menu-dynamics}).
On 2Wiki, the 3B run leads Search-R1, IGPO, and Graph-R1 throughout training and finishes with the highest F1 (Figure~\ref{fig:method-f1}).
Appendix~\ref{app:robust} reports numeric ranges and per-dataset traces.

\paragraph{Backbones and RL Algorithms.}
On HotpotQA under a matched budget, Qwen2.5-3B, Qwen3.5-4B, and Llama-3.2-3B all reach competitive F1 under the same GRPO recipe (Table~\ref{tab:robust-family}), so the method is not specific to one backbone family.
Fixing Qwen2.5-3B and the 120-step budget, GRPO, PPO, REINFORCE++, and DAPO \citep{schulman2017ppo,hu2025reinforcepp,shao2024deepseekmath,yu2025dapo} all rise stably, with GRPO slightly ahead of DAPO (Figure~\ref{fig:rl-estimators}; Appendix~\ref{app:robust}).
We use GRPO elsewhere unless stated otherwise.

\begin{table}[t]
\centering
\begin{minipage}[t]{0.46\columnwidth}
\centering
\footnotesize
\setlength{\tabcolsep}{2.4pt}
\renewcommand{\arraystretch}{1.12}
\begin{tabular}{@{}lccc@{}}
\toprule
Backbone & F1 & EM & G-E \\
\midrule
Qwen2.5-3B & 65.9 & 59.4 & 78.5 \\
Qwen3.5-4B & \textbf{67.5} & \textbf{61.2} & \textbf{79.8} \\
Llama-3.2-3B & 63.4 & 56.6 & 75.9 \\
\bottomrule
\end{tabular}
\caption{Backbone comparison.}
\label{tab:robust-family}
\end{minipage}\hfill
\begin{minipage}[t]{0.50\columnwidth}
\centering
\footnotesize
\setlength{\tabcolsep}{2.0pt}
\renewcommand{\arraystretch}{1.12}
\begin{tabular}{@{}l@{\hspace{0.4em}}c@{}}
\toprule
Item & Cost \\
\midrule
Train (8$\times$A100) & $\approx$170 GPU$\cdot$h \\
SNC / step & 51--69\,s (9--11\%) \\
Extra rollouts & 0 \\
\bottomrule
\end{tabular}
\caption{Training cost (Qwen2.5-3B).}
\label{tab:cost}
\end{minipage}
\end{table}
\setcounter{table}{6}% next table after forced Tab.~6

\subsection{Analysis of Interaction Efficiency (RQ4)}

As shown in Figure~\ref{fig:turns-length} and Tables~\ref{tab:cost}--\ref{tab:build}, we compare interaction cost, training overhead, and knowledge-construction cost.

\paragraph{More Turns, Shorter Responses.}
Late in training, Harness-G uses more turns than aligned Search-R1 and Graph-R1 ($3.6$ vs.\ $2.3$ and $2.9$) while keeping shorter response-side trajectories ($2.5$k vs.\ $4.3$k and $3.1$k tokens).
Structured navigation thus increases interaction without necessarily raising total response-side token cost.

\paragraph{Limited SNC Overhead.}
SNC requires no extra agent rollouts and adds only 9--11\% per-step wall-clock time, avoiding both Tree-GRPO's tree-expansion rollouts and GiGPO's specialized retroactive anchor-state grouping \citep{treegrpo2025,feng2025gigpo}; Table~\ref{tab:cost} reports about 170 GPU$\cdot$h for a full 3B run.

\paragraph{Cheap Construction, Low Query Cost.}
Table~\ref{tab:build} reports construction and per-query cost on 2Wiki under the protocol of \citet{luo2025graphr1}.
Because Harness-G builds the graph programmatically, API construction cost is \$0 (vs.\ \$2.81--\$4.14 for LLM-extracted graphs), with 0.12\,s per 1K corpus tokens and 5.1\,s per query at \$0 API cost, while matching or exceeding Graph-R1 F1 with a smaller 3B policy.
Measurement details are in Appendix~\ref{app:cost}.

\subsection{Cross-Dataset Generalization (RQ5)}

\begin{figure}[t]
\centering
\includegraphics[width=0.95\columnwidth]{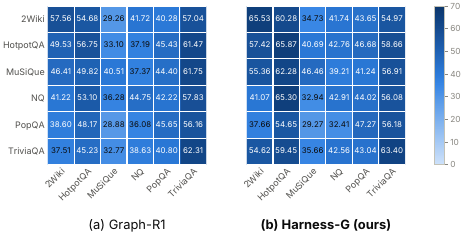}
\caption{F1 comparison across six datasets under O.O.D.\ cross-validation: (a)~Graph-R1; (b)~Harness-G.}
\label{fig:ood-f1}
\end{figure}

Figure~\ref{fig:ood-f1} compares Harness-G with Graph-R1 using Qwen2.5-3B under a train-on-one, evaluate-on-all-six setup; the 30 off-diagonal cells measure cross-dataset transfer.

\paragraph{Stronger O.O.D.\ Transfer.}
Harness-G wins 21 of 30 pairs and raises mean O.O.D.\ F1 from $44.10$ to $47.38$ ($+3.29$), with larger gains on multi-hop targets than on single-hop ones.

\paragraph{Robustness under Distribution Shift.}
The menu policy remains stable across source--target pairs rather than fitting a single dataset.
Additional cross-dataset and domain-transfer results are reported in Appendix~\ref{app:gen}.

% TODO: Add the wiki-18 open-corpus control once its final evaluation is available.

\section{Conclusion}

This work reframes a central design choice in RL search agents: the
retrieval action itself.
We show that free-form query generation can exhibit
retrieval-equivalence collapse, in which surface-level query diversity
masks nearly identical retrieved evidence and weakens group-relative
learning.
Harness-G replaces this aliased interface with finite evidence and entity
selections over a programmatically induced graph, while SNC exploits the
resulting action frontier to assign local and delayed credit without
additional agent rollouts.
Across six QA benchmarks, Harness-G improves average F1 over Graph-R1 by
3.98 points at 3B and 10.74 points at 1.5B.
Controlled ablations isolate gains from both the menu interface and SNC,
and cross-dataset evaluation indicates stronger transfer.
Together, these results support action-space design as a complementary
axis to reward design for training search agents.
Harness-G remains text-only; extending its structured actions and SNC to
multimodal evidence is a key next step.

\bibliography{references}

\clearpage
\appendix
\setcounter{secnumdepth}{2}

\section{Appendix Overview}
\label{app:overview}

This appendix collects details deferred from the main paper.
Appendix~\ref{app:data} specifies the six evaluation datasets, splits, and corpus preprocessing.
Appendix~\ref{app:baselines} describes the prompt-only and trained baselines in Table~\ref{tab:main}.
Appendix~\ref{app:metrics} defines EM, F1, generation evaluation, and answer normalization.
Appendix~\ref{app:impl} lists training hyperparameters and the full optimization algorithm.
Appendix~\ref{app:robust} expands training dynamics and robustness analyses.
Appendix~\ref{app:protocols} defines the retrieval-equivalence diagnostics in Figure~\ref{fig:diagnosis} and the zero-advantage diagnostics in Figure~\ref{fig:zadv}, and states the matched evaluation protocols used for interface and credit ablations.
Appendix~\ref{app:gen} reports additional multi-dataset and domain-transfer results.
Appendix~\ref{app:snc-alg} details per-trajectory SNC computation.
Appendix~\ref{app:action-screen} records the action-menu and environment-constraint screening that produced the final interface.
Appendix~\ref{app:filter} documents the invalid-target filter applied before menu rendering.
Appendix~\ref{app:cost} measures knowledge-construction and per-query cost.
Appendix~\ref{app:related} expands the related-work discussion.
Appendix~\ref{app:cases} analyzes several held-out trajectories, including failure cases.
Appendix~\ref{app:prompts} documents the environment protocol and prompt stack.

\section{Dataset Details}
\label{app:data}

Following Graph-R1 \citep{luo2025graphr1}, we evaluate Harness-G on six widely used RAG benchmarks covering both multi-hop and single-hop question answering.

\paragraph{Multi-hop datasets.}
\begin{itemize}
    \item \textbf{2WikiMultiHopQA (2Wiki)} \citep{ho20202wiki} contains compositional and comparison questions whose evidence is distributed across multiple Wikipedia pages.
    \item \textbf{HotpotQA} \citep{yang2018hotpotqa} includes bridge and comparison questions together with sentence-level supporting-fact annotations.
    \item \textbf{MuSiQue} \citep{trivedi2022musique} constructs multi-hop questions by composing single-hop questions, producing longer and less shortcut-prone reasoning chains.
\end{itemize}

\paragraph{Single-hop datasets.}
\begin{itemize}
    \item \textbf{Natural Questions (NQ)} \citep{kwiatkowski2019nq} is derived from real search-engine queries and pairs questions with Wikipedia evidence.
    \item \textbf{PopQA} \citep{mallen2023popqa} contains entity-centric open-domain questions spanning relations with different levels of entity popularity.
    \item \textbf{TriviaQA} \citep{joshi2017triviaqa} consists of trivia questions paired with distantly supervised evidence documents.
\end{itemize}

\paragraph{Splits and corpus preprocessing.}
Each dataset contains 5{,}120 training and 128 held-out questions; this split matches Graph-R1 for protocol comparability.
We report the final training checkpoint on the held-out split; no held-out question is used for checkpoint selection.
Because each held-out split is small, we emphasize large and consistent cross-dataset trends rather than isolated marginal differences.
Each dataset provides its own context corpus.
Following common agentic-RAG practice \citep{jin2025searchr1}, we segment each corpus into 1{,}200-token chunks with 100-token overlap, build one graph per dataset, and reuse both across all controlled variants.

\section{Baseline Details}
\label{app:baselines}

We organize the baselines in Table~\ref{tab:main} into prompt-only methods and Qwen2.5-based trained methods, following the grouping used by Graph-R1. The table additionally distinguishes the knowledge interface from the knowledge-construction regime. Here, \emph{LLM-based construction} means that corpus extraction or structural organization invokes a generative LLM; named-entity recognizers and dense encoders alone do not qualify.

\paragraph{Prompt-only and training-free methods.}
\begin{itemize}
    \item \textbf{Naive Generation} answers directly from the question without external retrieval.
    \item \textbf{Standard RAG} \citep{lewis2020rag} retrieves the top three E5 chunks \citep{wang2022e5} and conditions the generator on the retrieved text.
    \item \textbf{GraphRAG} \citep{edge2024graphrag} constructs an entity-centric graph and retrieves graph summaries for answer generation.
    \item \textbf{LightRAG} \citep{guo2024lightrag} uses a lightweight graph index with local and global retrieval.
    \item \textbf{PathRAG} \citep{chen2025pathrag} prunes and retrieves relational paths before generation.
    \item \textbf{HippoRAG2} \citep{gutierrez2025hipporag2} uses a non-parametric graph memory to support evidence retrieval.
    \item \textbf{HyperGraphRAG} \citep{luo2025hypergraphrag} uses hyperedges for higher-order relations and graph retrieval.
\end{itemize}
GraphRAG-family methods use GPT-4o-mini and are therefore marked as LLM-based.

\paragraph{Qwen2.5 reasoning and search methods.}
\begin{itemize}
    \item \textbf{Naive Generation} and \textbf{Standard RAG} use Qwen2.5-1.5B/3B as the generator, with no parameter update.
    \item \textbf{SFT} fine-tunes the backbone on question--answer examples without reinforcement learning.
    \item \textbf{R1} applies GRPO \citep{shao2024deepseekmath} to direct answer generation without retrieval.
    \item \textbf{Search-R1} \citep{jin2025searchr1} trains a multi-turn free-query search agent with outcome-reward GRPO.
    \item \textbf{IGPO} \citep{igpo2025} keeps the free-query interface of Search-R1 but densifies training with per-turn information-gain credit.
    \item \textbf{R1-Searcher} \citep{song2025r1searcher} uses staged reinforcement learning to acquire structured search behavior.
    \item \textbf{Graph-R1} \citep{luo2025graphr1} trains a multi-turn free-query agent over an LLM-constructed graph.
\end{itemize}

\paragraph{Harness-G and comparison scope.}
Harness-G is marked as programmatic, LLM-free knowledge construction because its graph uses sentence segmentation, entity recognition and normalization, and dense encoding without generative-LLM extraction or organization.
Baseline values in Table~\ref{tab:main} are published cross-stack Graph-R1 references; only the interface and credit variants in the RQ2 ablation analysis are matched, under the shared protocol in Appendix~\ref{app:protocols}.

\section{Evaluation Details}
\label{app:metrics}

Let $\hat y_i$ be the prediction for question $i$, $\mathcal{Y}_i$ its set of accepted answer aliases, and $N$ the number of evaluation questions.
We use the Graph-R1 evaluation suite and report all metrics on a $0$--$100$ scale.

\paragraph{Exact Match (EM).}
The normalization function $\operatorname{norm}(\cdot)$ lowercases text, removes articles and punctuation, and canonicalizes whitespace.
EM is the percentage of predictions matching any accepted alias:
\begin{equation}
\mathrm{EM}
=\frac{100}{N}\sum_{i=1}^{N}
\max_{y\in\mathcal{Y}_i}
\mathbb{I}\!\left[\operatorname{norm}(\hat y_i)=\operatorname{norm}(y)\right].
\end{equation}

\paragraph{Token F1.}
Let $\operatorname{tok}(\cdot)$ denote the normalized token multiset.
We take the maximum token-overlap F1 over gold aliases:
\begin{equation}
\mathrm{F1}
=\frac{100}{N}\sum_{i=1}^{N}
\max_{y\in\mathcal{Y}_i}
\frac{2\left|\operatorname{tok}(\hat y_i)\cap\operatorname{tok}(y)\right|}
{\left|\operatorname{tok}(\hat y_i)\right|+\left|\operatorname{tok}(y)\right|}.
\end{equation}

\paragraph{Retrieval Similarity (R-S).}
R-S measures semantic agreement between the retrieved evidence $K_i$ and gold evidence $K_i^\star$.
Using the Graph-R1 metric encoder $\operatorname{Enc}(\cdot)$,
\begin{equation}
\mathrm{R\mbox{-}S}
=\frac{100}{N}\sum_{i=1}^{N}
\cos\!\left(\operatorname{Enc}(K_i),\operatorname{Enc}(K_i^\star)\right).
\end{equation}

\paragraph{Generation Evaluation (G-E).}
Following Graph-R1, GPT-4o-mini scores each response from $0$ to $10$ on seven dimensions: comprehensiveness, knowledgeability, correctness, relevance, diversity, logical coherence, and factuality.
If $s_{i,d}$ is the score for dimension $d$, then
\begin{equation}
\mathrm{G\mbox{-}E}
=\frac{10}{7N}\sum_{i=1}^{N}\sum_{d=1}^{7}s_{i,d}.
\end{equation}
The factor of $10$ maps the result to the same $0$--$100$ scale as the other reported metrics.

\paragraph{Aggregation.}
Per-dataset columns in Table~\ref{tab:main} report F1 and G-E.
The Avg.\ columns are unweighted macro-averages over the six datasets for EM, F1, R-S, and G-E; they do not pool questions across datasets.

\section{Implementation Details}
\label{app:impl}

Table~\ref{tab:hyper} summarizes the canonical Qwen2.5-3B configuration used in the main experiments.
A single policy model both selects menu actions and generates the final answer.
We train on the verl framework \citep{sheng2024hybridflow} using eight NVIDIA A100 GPUs; under the batch configuration of Table~\ref{tab:hyper}, 120 optimizer steps correspond to approximately three passes over the training set.

\paragraph{Training objective and masking.}
Following Search-R1 \citep{jin2025searchr1}, environment-injected observation and retrieval tokens are excluded from the policy loss, so both GRPO and SNC update only model-generated tokens.
The frozen answerer used by SNC is the initial reference checkpoint, which remains fixed throughout training.

\paragraph{Numerical safeguards.}
Multi-turn agent RL is sensitive to unstable importance ratios, low-precision probability computation, and degenerate rollout groups \citep{wang2025ragen}.
We use dual-clip with $C{=}3.0$ to limit destructive updates from large negative advantages, compute policy probabilities from fp32 logits, and reject batches with non-finite gradients before the optimizer step.
For SNC, exponentials used in answer-probability scoring are clamped, and the batch-global credit scale is floored at $s_{\min}{=}5{\times}10^{-4}$.

\paragraph{Training loop.}
Algorithm~\ref{alg:training} summarizes rollout collection and policy optimization; Appendix~\ref{app:snc-alg} expands the per-trajectory SNC computation.

\begin{algorithm}[t]
\caption{Harness-G rollout and training.}
\label{alg:training}
\begin{algorithmic}[1]
\REQUIRE corpus $\mathcal{D}$, questions $\mathcal{Q}$, policy $\pi_\theta$, frozen answerer $g$, max turns $T$, frontier size $K_f$
\STATE Build the paragraph--sentence--entity graph $G$ from $\mathcal{D}$
\FOR{each training iteration}
  \FOR{each question $q$ in the batch, each of $N_g$ rollouts}
    \STATE $z_0 \leftarrow \mathrm{InitRetrieve}(q;G)$
    \FOR{$t=0,\ldots,T-1$}
      \STATE Render $o_t$ (committed evidence, visible sentences, menu $M_t$); sample $a_t\sim\pi_\theta(\cdot\mid o_t)$
      \STATE Read-only preview $a_t$ and frontier $\mathcal{F}_t$; execute $a_t$
      \IF{$a_t$ is \textsc{Answer}} \STATE \textbf{break} \ENDIF
    \ENDFOR
    \STATE Generate answer; compute $R^{\mathrm{out}}$; score previews with $g$; compute $r^{\mathrm{fr}}$, $r^{\mathrm{en}}$, and $r^{\mathrm{SNC}}=r^{\mathrm{fr}}+r^{\mathrm{en}}$ per step \eqref{eq:gain}--\eqref{eq:enable}; place on action spans
  \ENDFOR
  \STATE Update $\pi_\theta$ with the combined clipped objective \eqref{eq:combine}
\ENDFOR
\end{algorithmic}
\end{algorithm}

\begin{table}[!t]
\centering
\small
\begin{tabular}{@{}ll@{}}
\toprule
Parameter & Value \\
\midrule
\multicolumn{2}{@{}l}{\textit{Policy optimization}} \\
Backbone & Qwen2.5-3B-Instruct \\
GRPO group size $N_g$ & 8 \\
Batch size / mini-batch & 128 / 32 \\
Training steps & 120 ($\approx$3 epochs) \\
Learning rate & $5\times 10^{-7}$ \\
KL coefficient $\beta_{\mathrm{KL}}$ & 0.001 (low-variance KL) \\
Clip $\epsilon$ / dual-clip $C$ & 0.2 / 3.0 \\
Max prompt / response tokens & 8192 / 2048 \\
Max tool-response tokens & 4096 \\
Outcome reward & token-level answer F1 \\
Retrieved-token loss mask & enabled \\
\midrule
\multicolumn{2}{@{}l}{\textit{Environment and SNC}} \\
Max turns $T$ & 6 \\
Visible-sentence cap $K_s$ & 6 \\
\textsc{Lookup} candidate cap $K_e$ & 8 \\
Frontier size $K_f$ & 4 (type-stratified) \\
SNC weight $\lambda$ & 0.2 \\
Propagation discount $\gamma$ & 1.0 \\
Information-gain dead-zone & $10^{-4}$ \\
SNC clamp $c$ / scale floor $s_{\min}$ & 5.0 / $5{\times}10^{-4}$ \\
Frontier baseline & mean \\
Dependency rule & provenance \\
\textsc{Answer\_With} & enabled \\
Frozen answerer & initial reference checkpoint \\
\bottomrule
\end{tabular}
\caption{Hyperparameter settings for Harness-G (3B).}
\label{tab:hyper}
\end{table}

\paragraph{Graph construction.}
We first preserve document paragraphs and split them into sentences with abbreviation- and initial-safe rules.
Entity mentions are extracted with spaCy \texttt{en\_core\_web\_sm} for \textsc{person}, \textsc{org}, \textsc{gpe}, \textsc{loc}, \textsc{work\_of\_art}, \textsc{event}, \textsc{date}, \textsc{norp}, and \textsc{fac}.
Surface forms are canonicalized by whitespace and punctuation normalization, lowercasing, and junk filtering.
Paragraph--sentence edges preserve document context; sentence--entity edges record mentions; sentence--sentence edges connect adjacent sentences within a paragraph.
Canonical entities are linked to their top-five embedding neighbors when cosine similarity under \texttt{bge-large-en-v1.5} is at least $0.80$ \citep{xiao2023cpack}, and document titles are anchored to the first sentence.
No LLM is used for graph construction.
Each corpus graph is built once, cached, and reused across runs.

\section{Training Dynamics and Robustness Details}
\label{app:robust}

We expand the RQ3 stability and transfer analysis.

\paragraph{Stable optimization across datasets.}
Figure~\ref{fig:menu-dynamics} plots training-batch answer F1 and gradient norms for the six Qwen2.5-3B runs.
Across all six runs, training-batch F1 rises by $0.36$--$0.62$ without persistent late-stage collapse.
The recorded gradient norm is bounded by $12.7$ and is typically below $9$ late in training.

\paragraph{Backbone transfer.}
Scale comparisons at 1.5B and 3B appear in Table~\ref{tab:main}.
Here we hold the HotpotQA data and 120-step budget fixed and vary only the backbone family under the same GRPO recipe.
As shown in Table~\ref{tab:robust-family}, Qwen2.5-3B, Qwen3.5-4B, and Llama-3.2-3B obtain $63.4$--$67.5$ F1.
Qwen3.5-4B is best overall, while Llama-3.2-3B remains competitive within a $2.5$-point F1 band of the default, indicating that the method is not specific to one backbone family.

\paragraph{Compatibility with multiple RL algorithms.}
Figure~\ref{fig:rl-estimators} compares GRPO, PPO, REINFORCE++, and DAPO \citep{schulman2017ppo,hu2025reinforcepp,shao2024deepseekmath,yu2025dapo} with Qwen2.5-3B and the 120-step budget fixed.
Across all four settings, training-batch F1 rises rapidly and remains stable through optimization, with no persistent late-stage collapse.
GRPO performs best overall, only slightly ahead of DAPO, while REINFORCE++ and PPO exhibit similarly stable training.
These results show that Harness-G supports diverse RL algorithms rather than a particular optimizer; other experiments use GRPO.

\section{Additional Evaluation Protocols}
\label{app:protocols}

\paragraph{Retrieval-equivalence diagnostics.}
At training step \(s\), a rollout group \(\mathcal{G}_s(q)=\{\tau_i\}_{i=1}^{N_g}\) contains the \(N_g\) trajectories sampled for the same question \(q\).
If \(\mathcal{E}_{i,t}\) is the set of evidence returned at retrieval turn \(t\), the deduplicated accumulated evidence of trajectory \(\tau_i\) is
\begin{equation}
E(\tau_i)=\bigcup_{t=1}^{T_i}\mathcal{E}_{i,t}.
\end{equation}
Let \(\kappa(E(\tau_i))\) be its cluster assignment under the evidence-overlap clustering used in Figure~\ref{fig:diagnosis}.
Trajectories with the same assignment form a \emph{retrieval-equivalence class}, and the number of within-group classes is
\begin{equation}
N_{\mathrm{eq}}(\mathcal{G})
=\left|\left\{\kappa(E(\tau_i)):\tau_i\in\mathcal{G}\right\}\right|.
\end{equation}
Let \(\mathcal{B}_s\) be the rollout groups at step \(s\), and let \(\mathsf{Qry}(\mathcal{G})\) be the set of distinct query strings generated within group \(\mathcal{G}\).
The two curves in Figure~\ref{fig:diagnosis}a and the class count in Figure~\ref{fig:diagnosis}b are, respectively,
\begin{align}
D_{\mathrm{query}}(s)
&=\frac{1}{|\mathcal{B}_s|}\sum_{\mathcal{G}\in\mathcal{B}_s}
\mathbb{I}[|\mathsf{Qry}(\mathcal{G})|\geq2],\\
D_{\mathrm{ret}}(s)
&=\frac{1}{|\mathcal{B}_s|}\sum_{\mathcal{G}\in\mathcal{B}_s}
\mathbb{I}[N_{\mathrm{eq}}(\mathcal{G})\geq2],\\
\overline N_{\mathrm{eq}}(s)
&=\frac{1}{|\mathcal{B}_s|}\sum_{\mathcal{G}\in\mathcal{B}_s}
N_{\mathrm{eq}}(\mathcal{G}).
\end{align}
Here \(D_{\mathrm{query}}\) measures \emph{query-form diversity}, while \(D_{\mathrm{ret}}\) and \(\overline N_{\mathrm{eq}}\) measure \emph{retrieval-outcome diversity}.
We use \emph{retrieval-equivalence collapse} for the training-time movement of \(D_{\mathrm{ret}}\) toward zero and \(\overline N_{\mathrm{eq}}\) toward one, and \emph{illusory exploration} for the resulting regime in which \(D_{\mathrm{query}}\) remains high while retrieval-outcome diversity has collapsed.
All equivalence classes are therefore defined within a same-question rollout group rather than globally across unrelated questions.

\paragraph{Zero-advantage diagnostics.}
For the same-question rollout group $\mathcal{G}=\{\tau_i\}_{i=1}^{N_g}$ with outcome rewards $\{R_i\}$, we call $\mathcal{G}$ a \emph{zero-advantage group} when $N_g\geq2$ and the within-group reward standard deviation is below $10^{-6}$ (matching the GRPO baseline used in training).
We further stratify zero-advantage groups into \emph{all wrong} (every $R_i=0$) and \emph{all correct} (every $R_i>0$, i.e., every rollout obtains a strictly positive answer F1).
Figure~\ref{fig:zadv} reports the batch fraction of each class for Search-R1 and Harness-G on 2Wiki under matched Qwen2.5-3B / group-size-$8$ training.

\paragraph{Transition-matched interface comparison.}
At state $z_t$, the environment first constructs a latent feasible-target set $\mathcal{A}_t$ and deterministic transition $T(z_t,a)$ for every $a\in\mathcal{A}_t$.
The menu policy observes $\mathcal{A}_t$ as executable entries and emits a target id.
The matched free-query policy observes the same evidence state but not the executable ids; it emits an arbitrary string $u_t$, and a frozen resolver ranks the textual representations of targets in $\mathcal{A}_t$ with the shared encoder and deterministically selects $\rho(z_t,u_t)\in\mathcal{A}_t$.
Both arms then execute the same transition $T$.
They also share initial retrieval, visible and committed evidence, \textsc{Select}/\textsc{Answer}/\textsc{Answer\_With} semantics, target filtering and deduplication, candidate caps, top-$K$, maximum decision and retrieval-call budgets, evidence and token budgets, corpora, graph indices, initial checkpoints, questions, answer-F1 outcome reward, GRPO group size, batch size, 120-step budget, and retrieved-token loss masking.
Matched runs use the same data order and random seeds.
Consequently, the reachable next states and retrieval substrate are fixed; the intervention is whether a retrieval target is exposed for finite selection or reached through an open string-to-target mapping.

\paragraph{Factorial credit control.}
The outcome-only arms receive no additional step advantage.
The two IGPO arms \citep{igpo2025} use the same implementation, including policy-belief information gain, turn-level normalization, cumulative turn advantage, and reward scaling.
SNC is evaluated only on the action menu because frontier-relative credit requires an enumerable same-state action set; its gain over Menu+IGPO therefore measures the value of the structured frontier and provenance-based enablement terms beyond conventional trajectory-local information gain.

\paragraph{Cross-dataset generalization.}
Each source-dataset policy is evaluated on the held-out splits of all six target datasets under the protocol of Table~\ref{tab:main}.
For $D{=}6$ datasets, the reported O.O.D.\ mean over the 30 off-diagonal train--test pairs is
\begin{equation}
\overline{\mathrm{F1}}_{\mathrm{OOD}}
=\frac{1}{D(D-1)}\sum_{s=1}^{D}\sum_{\substack{t=1\\t\neq s}}^{D}\mathrm{F1}_{s,t}.
\end{equation}

\section{Additional Generalization Results}
\label{app:gen}

This appendix extends the RQ5 cross-dataset study.
Under the train-on-one, evaluate-on-all-six protocol of Figure~\ref{fig:ood-f1}, Harness-G outperforms Graph-R1 in 21 of 30 off-diagonal pairs and raises mean O.O.D.\ F1 from $44.10$ to $47.38$ ($+3.29$).
Averaged over the five off-diagonal sources per target, the margin is $+6.45$ F1 on the multi-hop targets 2Wiki, HotpotQA, and MuSiQue, versus $+0.12$ on the single-hop targets NQ, PopQA, and TriviaQA.
This contrast is consistent with the action menu being most useful when transfer requires multi-hop evidence composition.
We next report multi-dataset training regimes and zero-shot transfer to specialized non-Wikipedia domains, following axes evaluated by \citet{luo2025graphr1}.

\paragraph{Multi-dataset training regimes.}
We compare three regimes at the 3B scale.
Under \emph{I.I.D.}\ training, each policy is trained and evaluated on the same dataset, as in Table~\ref{tab:main}.
Under \emph{single-O.O.D.}\ training, the policy is trained on one source dataset and evaluated on a different target; per-target values average over the five non-matching sources and correspond to the off-diagonal cells of Figure~\ref{fig:ood-f1}.
Under \emph{combined} training, the six training pools are merged and uniformly subsampled to one sixth of the union, matching the single-dataset training volume.
Table~\ref{tab:regimes} reports EM and F1 for Search-R1, Graph-R1, and Harness-G.

\begin{table*}[t]
\centering
\footnotesize
\setlength{\tabcolsep}{3.2pt}
\renewcommand{\arraystretch}{1.12}
\begin{tabular}{@{}l*{14}{c}@{}}
\toprule
& \multicolumn{2}{c}{2Wiki} & \multicolumn{2}{c}{HotpotQA} & \multicolumn{2}{c}{MuSiQue} & \multicolumn{2}{c}{NQ} & \multicolumn{2}{c}{PopQA} & \multicolumn{2}{c}{TriviaQA} & \multicolumn{2}{c}{Avg.} \\
\cmidrule(lr){2-3}\cmidrule(lr){4-5}\cmidrule(lr){6-7}\cmidrule(lr){8-9}\cmidrule(lr){10-11}\cmidrule(lr){12-13}\cmidrule(l){14-15}
Method & EM & F1 & EM & F1 & EM & F1 & EM & F1 & EM & F1 & EM & F1 & EM & F1 \\
\midrule
\multicolumn{15}{@{}l}{\textit{I.I.D.: trained on the target dataset}} \\
\addlinespace[1pt]
Search-R1 & 31.25 & 38.04 & 38.28 & 43.84 & 3.91 & 7.65 & 24.22 & 37.96 & 33.59 & 38.67 & 40.62 & 47.99 & 28.65 & 35.69 \\
Graph-R1 & 50.00 & 57.56 & 50.78 & 56.75 & 32.81 & 40.51 & 30.47 & \textbf{44.75} & 37.50 & 45.65 & 53.13 & 62.31 & 42.45 & 51.26 \\
\textbf{Harness-G} & \textbf{59.38} & \textbf{65.53} & \textbf{59.38} & \textbf{65.87} & \textbf{35.16} & \textbf{46.46} & 32.81 & 42.91 & \textbf{42.19} & \textbf{47.27} & \textbf{53.91} & \textbf{63.40} & \textbf{47.14} & \textbf{55.24} \\
\midrule
\multicolumn{15}{@{}l}{\textit{Single O.O.D.: trained on one non-target dataset (averaged over the five sources)}} \\
\addlinespace[1pt]
Search-R1 & -- & 24.30 & -- & 29.40 & -- & 5.64 & -- & 25.46 & -- & 26.59 & -- & 40.29 & -- & 25.28 \\
Graph-R1 & -- & 42.65 & -- & 50.20 & -- & 32.06 & -- & 38.20 & -- & 42.63 & -- & \textbf{58.85} & -- & 44.10 \\
\textbf{Harness-G} & -- & \textbf{49.23} & -- & \textbf{60.39} & -- & \textbf{34.66} & -- & \textbf{39.74} & -- & \textbf{43.73} & -- & 56.53 & -- & \textbf{47.38} \\
\midrule
\multicolumn{15}{@{}l}{\textit{Combined: trained on the merged pool subsampled to matched volume}} \\
\addlinespace[1pt]
Search-R1 & 24.22 & 30.15 & 34.38 & 39.82 & 8.59 & 13.97 & 23.44 & 36.21 & 34.38 & 38.85 & 43.75 & 51.38 & 28.13 & 35.06 \\
Graph-R1 & 43.75 & 50.84 & 48.44 & 55.61 & 30.47 & 37.15 & 28.13 & 40.72 & 39.06 & 46.18 & \textbf{51.56} & \textbf{59.73} & 40.24 & 48.37 \\
\textbf{Harness-G} & \textbf{53.91} & \textbf{60.48} & \textbf{55.47} & \textbf{62.95} & \textbf{34.38} & \textbf{45.12} & \textbf{31.25} & \textbf{41.86} & \textbf{41.41} & \textbf{46.92} & 50.78 & 59.41 & \textbf{44.53} & \textbf{52.79} \\
\bottomrule
\end{tabular}
\caption{Three training regimes on six datasets (Qwen2.5-3B). Best in each regime in \textbf{bold}.}
\label{tab:regimes}
\end{table*}

Harness-G attains the highest average F1 under all three regimes.
The regime ordering reported for free-query agents---I.I.D.\ best, combined second, single-O.O.D.\ worst---also holds for Harness-G, but with smaller gaps between regimes, indicating that menu navigation over a corpus-induced graph transfers across training distributions rather than fitting a single dataset.

\paragraph{Zero-shot transfer to specialized domains.}
We evaluate the combined-trained 3B policies zero-shot on the five specialized-domain benchmarks introduced by HyperGraphRAG \citep{luo2025hypergraphrag}---Medicine, Agriculture, Computer Science (CS), Legal, and a mixed-domain split (Mix)---with no training data from these domains.
For each domain corpus, Harness-G induces its paragraph--sentence--entity graph with the same programmatic pipeline as in Appendix~\ref{app:impl}; because construction invokes no generative LLM, extending to a new domain incurs no extraction cost.
Table~\ref{tab:domains} reports the results.

\begin{table*}[t]
\centering
\footnotesize
\setlength{\tabcolsep}{3.6pt}
\renewcommand{\arraystretch}{1.12}
\begin{tabular}{@{}l*{12}{c}@{}}
\toprule
& \multicolumn{2}{c}{Medicine} & \multicolumn{2}{c}{Agriculture} & \multicolumn{2}{c}{CS} & \multicolumn{2}{c}{Legal} & \multicolumn{2}{c}{Mix} & \multicolumn{2}{c}{Avg.} \\
\cmidrule(lr){2-3}\cmidrule(lr){4-5}\cmidrule(lr){6-7}\cmidrule(lr){8-9}\cmidrule(lr){10-11}\cmidrule(l){12-13}
Method & EM & F1 & EM & F1 & EM & F1 & EM & F1 & EM & F1 & EM & F1 \\
\midrule
\multicolumn{13}{@{}l}{\textit{GPT-4o-mini (prompt-only)}} \\
\addlinespace[1pt]
Naive generation & -- & 12.89 & -- & 12.74 & -- & 18.65 & -- & 21.64 & -- & 16.93 & -- & 16.57 \\
Standard RAG & -- & 27.90 & -- & 27.43 & -- & 28.93 & -- & 37.34 & -- & 43.20 & -- & 32.96 \\
GraphRAG & -- & 17.60 & -- & 21.28 & -- & 23.33 & -- & 30.11 & -- & 19.27 & -- & 22.32 \\
LightRAG & -- & 12.79 & -- & 18.24 & -- & 22.72 & -- & 31.64 & -- & 27.03 & -- & 22.48 \\
PathRAG & -- & 14.94 & -- & 21.30 & -- & 26.73 & -- & 31.29 & -- & 37.07 & -- & 26.27 \\
HippoRAG2 & -- & 21.34 & -- & 12.63 & -- & 17.34 & -- & 18.53 & -- & 21.53 & -- & 18.27 \\
HyperGraphRAG & -- & 35.35 & -- & 33.89 & -- & 31.30 & -- & \underline{43.81} & -- & 48.71 & -- & 38.61 \\
\midrule
\multicolumn{13}{@{}l}{\textit{Qwen2.5-3B-Instruct (combined training, zero-shot)}} \\
\addlinespace[1pt]
Search-R1 & 12.50 & 20.96 & 12.11 & 18.42 & 16.80 & 24.58 & 15.23 & 25.17 & 22.27 & 30.45 & 15.78 & 23.92 \\
Graph-R1 & \underline{23.83} & \underline{36.71} & \underline{26.17} & \underline{38.05} & \underline{26.56} & \underline{36.92} & \underline{22.66} & 34.18 & \underline{37.89} & \underline{49.84} & \underline{27.42} & \underline{39.14} \\
\textbf{Harness-G} & \textbf{25.39} & \textbf{39.28} & \textbf{28.52} & \textbf{41.17} & \textbf{28.91} & \textbf{39.64} & \textbf{24.61} & 37.52 & \textbf{40.63} & \textbf{53.91} & \textbf{29.61} & \textbf{42.30} \\
\bottomrule
\end{tabular}
\caption{Domain-wise results on five specialized datasets. Best in \textbf{bold}, second \underline{underlined}.}
\label{tab:domains}
\end{table*}

Harness-G obtains the best average EM and F1 in this zero-shot setting, exceeding Graph-R1 by 3.16 average F1 and the strongest prompt-only reference, HyperGraphRAG, by 3.69.
The gains are consistent across knowledge-intensive verticals such as Medicine and CS, indicating that structured menu navigation, rather than domain-specific supervision, drives the transfer.

\section{SNC Computation per Trajectory}
\label{app:snc-alg}

Algorithm~\ref{alg:snc} expands the SNC computation for one completed trajectory.
The computation has two stages: a same-state comparison against feasible frontier actions and reverse credit propagation over the provenance graph.
All previews are read-only and use the same deterministic retrieval operators as actual environment transitions.

\paragraph{Answerer scoring.}
The frozen answerer computes the length-normalized teacher-forced gold-answer probability
\begin{equation}
g(O)=\exp\Big(\tfrac{1}{|y^*|}\textstyle\sum_{j}\log P_{\bar\theta}(y^*_j\mid q,O,y^*_{<j})\Big),
\end{equation}
with stop-gradient parameters $\bar\theta$, maximized over the normalized gold-answer alias set from the dataset annotations.

\begin{algorithm}[t]
\caption{SNC credit computation for one trajectory.}
\label{alg:snc}
\begin{algorithmic}[1]
\REQUIRE trajectory $\tau=\{(z_t,a_t,M_t)\}_{t=1}^{T}$, $\mathrm{Preview}$, answerer $g$
\FOR{each step $t$}
  \STATE $O_t \leftarrow$ observed evidence before $a_t$
  \STATE $(\widetilde{U}_t, \widetilde{V}_t) \leftarrow \mathrm{Preview}(z_t,a_t)$
  \STATE $p_t(a_t)\leftarrow g(O_t\cup\widetilde{U}_t)-g(O_t)$
  \IF{$a_t$ acquires information}
    \STATE Draw frontier $\mathcal{F}_t\subseteq M_t\setminus\{a_t\}$ of information-acquisition actions (type-stratified, top-$K_f$)
    \FOR{$a\in\mathcal{F}_t$}
      \STATE $p_t(a)\leftarrow g(O_t\cup\mathrm{Preview}(z_t,a).U)-g(O_t)$
    \ENDFOR
    \STATE If $\mathcal{F}_t\ne\varnothing$, set $r^{\mathrm{fr}}_t\leftarrow p_t(a_t)-\mathrm{mean}_{a\in\mathcal{F}_t}p_t(a)$; otherwise set it to $0$
  \ELSE
    \STATE $r^{\mathrm{fr}}_t\leftarrow 0$
  \ENDIF
\ENDFOR
\STATE Build provenance DAG $\mathcal{D}_\tau$ using the latest direct producer
\STATE Initialize $r^{\mathrm{en}}_t\leftarrow 0$ for all $t$
\FOR{each step $t$ in reverse topological order}
  \STATE $R_t\leftarrow p_t(a_t)+\gamma\, r^{\mathrm{en}}_t$
  \STATE Distribute $R_t/|\mathrm{Pred}(t)|$ to the direct producers
\ENDFOR
\FOR{each step $t$}
  \STATE $r^{\mathrm{SNC}}_t\leftarrow r^{\mathrm{fr}}_t+r^{\mathrm{en}}_t$
\ENDFOR
\end{algorithmic}
\end{algorithm}

\paragraph{Frontier construction.}
At an information-acquisition step, the frontier contains at most $K_f{=}4$ other information-acquisition actions, deduplicated by resulting scorer context and stratified among eligible action types.
Commit and terminal steps use an empty frontier and $r_t^{\mathrm{fr}}=0$.

\paragraph{Provenance and token placement.}
The dependency graph records the latest step that produced each sentence or entity later consumed by another action.
Reverse propagation assigns delayed credit to those direct producers without crediting unrelated earlier steps.
The resulting SNC value is placed only on the generated token span corresponding to the selected action; observation tokens and tool responses receive zero SNC credit.

\paragraph{Computation.}
Answerer calls for the selected action and frontier alternatives are batched and cached within a step.
SNC introduces no additional policy rollout: it reuses the realized trajectory and adds only read-only index queries plus frozen-answerer scoring.
The measured overhead is reported in Table~\ref{tab:cost} of the main paper.

\section{Action-Space and Constraint Screening}
\label{app:action-screen}

The final Harness-G interface exposes four policy actions---\textsc{Select}, \textsc{Lookup}, \textsc{Answer\_With}, and \textsc{Answer}---together with environment-side constraints that never enter the menu.
This appendix records how that design was reached.
We separate \emph{action-menu} choices from \emph{environment constraints}, report all numbers on \textbf{2WikiMultiHopQA} with a Qwen2.5-3B-Instruct backbone under a matched training budget, and use F1, EM, and mean interaction turns as complementary metrics.
Analysis is given in the text; the tables list only configurations and scores.
The final row of each table coincides with the full Menu+SNC setting reported in the main experiments.

\subsection{Action-menu evolution}
\label{app:action-menu}

\paragraph{Action inventory.}
We first fix the meaning of every action name that appears in Table~\ref{tab:action-screen}.
Final-menu operators match those defined in \emph{The Menu Environment}:
\textsc{Select}$(s)$ commits a visible sentence $s\in U_t$ into the evidence set $C_t$;
\textsc{Lookup}$(e)$ treats entity $e$ as an information target and retrieves candidate sentences about $e$, with the retrieval query constructed by the environment rather than authored by the policy;
\textsc{Answer} stops retrieval and generates $\hat y$ from $C_t$;
\textsc{Answer\_With}$(s_1,\ldots,s_k)$ commits one or more visible sentences and terminates in a single step.
Legacy operators that appear only during screening are defined as follows.
\textsc{Expand\_Entity}$(e)$ expands from a mentioned entity by retrieving additional sentences that mention $e$ (or its local graph neighbors), exposing a neighborhood-expansion mechanism to the policy.
\textsc{Bridge\_Entity}$(e)$ is a second entity-centric hop intended to surface cross-document bridge entities for multi-hop questions; operationally it also issues an entity-conditioned retrieval, so it competes with \textsc{Expand\_Entity} for the same credit.
\textsc{Open\_Context}$(s)$ reveals adjacent sentences of an already visible sentence $s$ without changing the committed evidence $C_t$.
\textsc{Rewrite\_Query}$(u)$ lets the policy emit a free-form string $u$ that is sent to the retriever, reintroducing open-text query generation inside an otherwise typed menu.
\textsc{Stop} is the legacy name for episode termination and is renamed to \textsc{Answer} without changing its semantics.

\paragraph{Screening path.}
Early navigation interfaces mixed \emph{what} information to pursue with \emph{how} to retrieve it in a single menu.
The policy had to adopt evidence sentences, choose among mechanism-level hops such as \textsc{Expand\_Entity} and \textsc{Bridge\_Entity}, optionally open local context, rewrite a free-form rescue query, and stop.
Trajectory logs from this legacy menu are highly unbalanced: \textsc{Select} and \textsc{Expand\_Entity} dominate executed steps, \textsc{Bridge\_Entity} varies widely across runs, \textsc{Open\_Context} is offered often but almost never taken, and \textsc{Rewrite\_Query} either collapses into a free-query escape hatch or is disabled by menu repair.
The resulting action space is long, hard to credit, and unstable under group-relative training.

We therefore redesign the menu so that the policy only chooses information \emph{targets}, while retrieval mechanisms are executed inside the environment.
Table~\ref{tab:action-screen} summarizes that progression on 2Wiki.
L0 is the Search-R1 free-query interface on the same backbone and dataset: the policy writes a natural-language retrieval query each turn and answers from retrieved passages.
L1 replaces free-form queries with the full legacy typed menu on the corpus-induced graph.
Despite exposing structured operators, L1 underperforms L0: the menu forces the policy to choose among redundant mechanisms, near-dead local-context actions, and a rewrite escape hatch, so group-relative credit is diluted and multi-hop F1 drops below free-query Search-R1.
Removing \textsc{Rewrite\_Query} closes that free-query back door without changing the rest of the typed menu.
Dropping \textsc{Open\_Context} removes a near-dead action whose selection rate stays far below its offer rate; local sentence context is instead surfaced through neighborhood retrieval after a target hop.
Merging \textsc{Expand\_Entity} and \textsc{Bridge\_Entity} into a single \textsc{Lookup} eliminates a forced mechanism choice between two interchangeable entity-centric hops and yields the first large jump that surpasses L0.
Renaming \textsc{Stop} to \textsc{Answer} is purely notational.
Adding \textsc{Answer\_With} reduces termination friction: when a visible sentence already suffices, the policy can commit it and finish in one step instead of \textsc{Select} then \textsc{Answer}.
Internalizing the \textsc{Lookup} query---concatenating the question with committed evidence inside the environment---further stabilizes multi-hop retrieval because the policy no longer authors an open need string.
The last step keeps this four-action menu fixed and adds Structured Non-myopic Credit (SNC); the gain relative to the preceding row is therefore credit assignment rather than another action type.
The final Menu+SNC row matches the main 2Wiki result.

\begin{table*}[t]
\centering
\small
\setlength{\tabcolsep}{4pt}
\renewcommand{\arraystretch}{1.12}
\begin{tabular}{@{}lp{0.58\textwidth}rrr@{}}
\toprule
Stage & Action menu & F1 & EM & Turns \\
\midrule
L0 (Search-R1) & free-form retrieval query + answer
& 38.0 & 31.3 & 2.3 \\
L1 full legacy menu & \textsc{Select}, \textsc{Open\_Context}, \textsc{Bridge\_Entity}, \textsc{Expand\_Entity}, \textsc{Rewrite\_Query}, \textsc{Stop}
& 28.6 & 21.4 & 5.2 \\
L1a drop rewrite & \textsc{Select}, \textsc{Open\_Context}, \textsc{Bridge\_Entity}, \textsc{Expand\_Entity}, \textsc{Stop}
& 30.1 & 22.8 & 5.1 \\
L1b drop open & \textsc{Select}, \textsc{Bridge\_Entity}, \textsc{Expand\_Entity}, \textsc{Stop}
& 30.5 & 23.1 & 5.0 \\
L2 merge mechanisms & \textsc{Select}, \textsc{Lookup}, \textsc{Stop}
& 41.8 & 33.2 & 4.6 \\
L3 rename stop & \textsc{Select}, \textsc{Lookup}, \textsc{Answer}
& 42.0 & 33.5 & 4.6 \\
L3h + answer-with & \textsc{Select}, \textsc{Lookup}, \textsc{Answer\_With}, \textsc{Answer}
& 54.6 & 45.2 & 3.9 \\
L3q internalized lookup & same four actions; environment builds the lookup query
& 61.2 & 51.8 & 3.5 \\
\midrule
L$^\star$ final menu + SNC & \textsc{Select}, \textsc{Lookup}, \textsc{Answer\_With}, \textsc{Answer}
& \textbf{65.5} & \textbf{59.4} & 3.6 \\
\bottomrule
\end{tabular}
\caption{Action-menu screening on 2WikiMultiHopQA (Qwen2.5-3B). L0 is Search-R1 free-query; later rows are menu variants on the corpus-induced graph. F1 and EM in percent; Turns is the mean number of environment interactions per question.}
\label{tab:action-screen}
\end{table*}

\subsection{Environment constraints}
\label{app:action-constraints}

In parallel with menu screening, two environment rules restrict \emph{which} entities may serve as navigation anchors and \emph{whether} a lookup may spin on a repeated key.
Neither rule is an action: the policy never selects ``filter'' or ``dedup''; both are applied before or during menu construction.
Without them, entity-centric hops still land on weak anchors---dates, cardinals, nationality adjectives, and fragmented person surfaces such as name-plus-isolated-initial---or repeat the same \textsc{Lookup} with no new evidence.
Table~\ref{tab:constraint-screen} isolates these rules on the fixed four-action menu (\textsc{Select}, \textsc{Lookup}, \textsc{Answer\_With}, \textsc{Answer}) before SNC is enabled, then restores full SNC in the last row.

Junk and bad-target filtering removes typed literals and malformed surfaces from the \textsc{Lookup} menu while leaving the mention visible in sentence text for \textsc{Select} and answering; Appendix~\ref{app:filter} lists the excluded labels and surface heuristics.
Lookup deduplication forbids re-offering the same entity (and the same environment-built need key) after it has already been looked up in the episode, which cuts no-op turns more than it moves EM.
Gains from constraints are smaller than those from mechanism merge or \textsc{Answer\_With}, but they stabilize turns and keep the feasible frontier semantically meaningful.
The full configuration---filtered, deduplicated menu plus SNC---is the main Harness-G interface.

\begin{table}[t]
\centering
\small
\setlength{\tabcolsep}{3.5pt}
\renewcommand{\arraystretch}{1.12}
\begin{tabular}{@{}lrrr@{}}
\toprule
Configuration & F1 & EM & Turns \\
\midrule
Four-action menu, no extra constraints
& 56.8 & 45.0 & 4.2 \\
+ junk / bad-target filter
& 60.5 & 48.9 & 3.9 \\
+ lookup dedup (on top of the filter)
& 61.8 & 50.1 & 3.7 \\
\midrule
Filter + dedup + SNC (full setting)
& \textbf{65.5} & \textbf{59.4} & 3.6 \\
\bottomrule
\end{tabular}
\caption{Environment-constraint screening on 2WikiMultiHopQA with the final four-action menu fixed (Qwen2.5-3B). Constraints are not policy actions.}
\label{tab:constraint-screen}
\end{table}

\section{Invalid-Target Filter}
\label{app:filter}

The invalid-target filter is the concrete realization of the junk / bad-target constraint in Appendix~\ref{app:action-constraints}.
It removes actions that are syntactically executable but rarely define a useful retrieval hop.
Filtering changes only the feasible \textsc{Lookup} menu; the original mention remains visible in sentence text and can still support answer generation.
We exclude three groups:
\begin{itemize}
    \item typed mentions with NER labels \textsc{date}, \textsc{time}, \textsc{cardinal}, \textsc{ordinal}, \textsc{quantity}, \textsc{percent}, \textsc{money}, and \textsc{norp};
    \item malformed surfaces, including pure numbers, single characters, quote-containing fragments, truncated aliases, and name-plus-isolated-initial fragments; and
    \item nationality or ethnic adjectives from an explicit lexicon (e.g., \emph{american}, \emph{british}, and \emph{european}) for cases where rule-based extraction provides no reliable type.
\end{itemize}
This filtering is applied before menu rendering, so the policy is not asked to learn penalties for actions that the environment can identify as invalid deterministically.

\section{Knowledge-Construction and Query Cost}
\label{app:cost}

Table~\ref{tab:build} in the main text compares construction and per-query costs on 2Wiki under the metrics of \citet{luo2025graphr1}: time per 1K corpus tokens (TP1KT), API cost per 1M corpus tokens (CP1MT), graph size, time per query (TPQ), and API cost per 1K queries (CP1KQ).
Because Harness-G's construction pipeline invokes no generative LLM, construction time is dominated by sentence splitting, entity recognition, and dense encoding.
On a single A100, Harness-G processed 3{,}247{,}308 corpus tokens in 376.83\,s (0.116\,s TP1KT).
Query latency was measured with eight data-parallel A100 workers after one warm-up query per worker (model initialization excluded): 81.43\,s for 16 held-out questions (5.09\,s TPQ).
Training-time compute for the 3B configuration is reported in Table~\ref{tab:cost}.
SNC adds no agent rollouts and 9--11\% wall-clock time per training step (51--69\,s); it avoids Tree-GRPO's tree-expansion rollouts and GiGPO's specialized retroactive anchor-state grouping \citep{treegrpo2025,feng2025gigpo}.
The overhead scales with the bounded preview frontier: GPU hours cover rollout generation and policy optimization, while SNC accounts for read-only previews and frozen-answerer scoring of the selected action and its alternatives.
SNC is training-only and adds no model calls at inference.
A full 3B run costs about 170 GPU$\cdot$h on 8$\times$A100.

\section{Extended Related Work}
\label{app:related}

This appendix expands the main paper's related-work discussion, organizing prior work by retrieval \emph{action}, knowledge \emph{structure}, and training \emph{signal} before contrasting Harness-G along each axis.

\paragraph{Free-form RL search agents.}
RL-based query reformulation already optimized document recall or answer quality through natural-language rewrites \citep{nogueira2017query,buck2018activeqa}.
WebGPT exposed typed browser commands around a free-form search operation \citep{nakano2021webgpt}, and ReAct generalized the pattern of interleaving reasoning traces with external actions \citep{yao2023react}.
IRCoT and FLARE interleave generation with repeated retrieval \citep{trivedi2023ircot,jiang2023flare}; Self-RAG \citep{asai2024selfrag} and related adaptive RAG controllers further learn \emph{when} and \emph{what} to retrieve under reflection or complexity-aware policies, but typically use prompting or supervised objectives rather than multi-turn group-relative RL.
Search-R1 \citep{jin2025searchr1} made outcome-reward RL the default for multi-turn search: the policy emits free-form queries, a black-box retriever returns passages, and the final answer F1 (or EM) is the sole trajectory return.
R1-Searcher \citep{song2025r1searcher} and ReSearch \citep{chen2025research} strengthen the same free-query interface with cold-start or reasoning-oriented RL recipes; DeepResearcher \citep{zheng2025deepresearcher} and WebThinker \citep{li2025webthinker} push toward live web and long-horizon research settings; Search-o1 \citep{li2025searcho1} couples agentic search with large reasoning models at inference time; ZeroSearch \citep{sun2025zerosearch} and related work reduce live-search cost by simulating retrieval during training.
Across these free-query methods, the retrieval target remains expressed as an unrestricted natural-language string.
Harness-G retains outcome-supervised multi-turn search but replaces free strings with a discrete, environment-managed action menu.

\paragraph{Dense process rewards and multi-turn credit.}
Sparse terminal rewards make multi-turn agent RL brittle.
Classical methods redistribute delayed returns or infer past-action credit in hindsight \citep{arjona2019rudder,harutyunyan2019hca}.
RAGEN \citep{wang2025ragen} diagnoses self-evolution pathologies under multi-turn RL, including vanishing within-group advantages.
IGPO \citep{igpo2025} densifies search training with per-turn information gain measured by gold-answer probability under a frozen answerer; StepSearch \citep{wang2025stepsearch} applies step-wise PPO with intermediate retrieval supervision; ReasonRAG \citep{zhang2025reasonrag} studies process versus outcome rewards for agentic RAG; CriticSearch \citep{zhang2026criticsearch} retrospectively scores turns with a frozen critic; Tree-GRPO \citep{treegrpo2025} creates step-level contrast through tree expansion with additional rollouts, whereas GiGPO \citep{feng2025gigpo} retroactively groups actions from repeated anchor states in already collected trajectories, without additional rollouts but with specialized grouping.
Turn-level advantage estimators for general multi-turn agents follow a similar densification logic \citep{wei2025turnlevel,li2025turnppo}.
These methods improve \emph{how} credit is estimated inside the free-query (or free-tool) space; they do not remove retrieval-equivalence classes induced by linguistic aliasing.
SNC is complementary but interface-dependent: only because the menu enumerates and previews feasible alternatives can same-state, frontier-relative step credit be computed as pure environment-side bookkeeping, without tree sampling.

\paragraph{Graph-structured retrieval.}
Chunk RAG \citep{lewis2020rag} flattens evidence; GraphRAG \citep{edge2024graphrag}, LightRAG \citep{guo2024lightrag}, PathRAG \citep{chen2025pathrag}, HyperGraphRAG \citep{luo2025hypergraphrag}, and HippoRAG / HippoRAG~2 \citep{gutierrez2024hipporag,gutierrez2025hipporag2} reorganize corpora into entity-, path-, or community-structured indices to support multi-hop aggregation.
Most pipelines invoke generative LLMs for OpenIE-style triple extraction, relation labeling, or community summaries, which raises construction cost and injects extraction noise.
HippoRAG~2 is structurally close to our graph---passages linked through entities---but is built by GPT-4o-mini OpenIE and used for one-shot retrieval, not interactive RL.
Graph-R1 \citep{luo2025graphr1} is the closest agentic system: it trains multi-turn interaction against a GraphRAG retriever with end-to-end RL.
Its graph is still LLM-extracted, and the policy still \emph{generates} free-form queries against that graph tool.
Harness-G shares the ``retrieval as interactive decision process'' view, but (i)~builds a relation-free paragraph--sentence--entity graph programmatically (no generative LLM in construction) and (ii)~exposes only finite, deduplicated, environment-validated menu actions, so query construction and state updates are deterministic environment operations.

\paragraph{Discrete navigation on knowledge graphs.}
Path-based KGQA long treated multi-hop answering as sequential decision making.
MINERVA \citep{das2018minerva} learns RL policies that walk curated knowledge bases by choosing relations; Think-on-Graph and ToG~2.0 \citep{sun2024tog,ma2024tog2} prompt LLMs to explore entity--relation frontiers; KG-R1 \citep{kgr12025} and Search-on-Graph-R1 \citep{sun2026sogr1} train RL agents to search structured graphs more efficiently.
These systems enjoy finiteness and (often) verifiability because the schema and triple interface are given.
They do not address open-text corpora without a manually defined KG, nor do they turn a cheaply induced evidence graph into a previewable RL harness for free-form RAG agents.
Harness-G imports the discrete-navigation action regime into open documents: sentences are the atomic evidence units, entities are bridge nodes without relation labels, and the menu---not the triple language---is the policy interface.

\paragraph{Summary of the gap.}
Prior work either densifies rewards under free-form queries, structures knowledge for one-shot or free-query retrieval, or discretizes actions only on curated KGs.
Harness-G instead combines a programmatic open-corpus graph with a finite, verifiable, previewable menu, making distinct exploration and non-myopic credit properties of the environment.

\section{Illustrative Trajectories}
\label{app:cases}

This appendix provides qualitative evidence that complements the quantitative RQ2 interface and step-credit ablations.
We analyze several held-out trajectories of the Menu+SNC Qwen2.5-3B policy under the evaluation protocol of Table~\ref{tab:main}, covering bridge retrieval, film-year comparison, multi-hop composition, birth-date comparison, and boolean country comparison (Figures~\ref{fig:case-bridge}--\ref{fig:case-country}).
Each menu figure records (i)~the question and gold answer, (ii)~the opening and post-commit menus with chosen vs.\ rejected targets, and (iii)~the final committed set $C_T$ that conditions the answer.
Figure~\ref{fig:case-freequery} contrasts the same bridge question under a free-query interface.
Figures~\ref{fig:case-snc-fr}--\ref{fig:case-snc-myopic} unpack SNC on that bridge trajectory.
Figures~\ref{fig:case-fail-attr}--\ref{fig:case-fail-coref} analyze additional held-out HotpotQA trajectories of the same Menu+SNC Qwen2.5-3B checkpoint where the final answer is incorrect.
Together, the success cases make three claims \emph{operational}: (1)~the menu reparameterizes retrieval into finite typed decisions; (2)~that reparameterization yields early termination and distractor rejection when evidence is already visible; and (3)~SNC turns the same menu into non-myopic step credit that outcome-only and myopic process rewards cannot supply.
The failure cases complement those illustrations by showing how the same typed action log behaves on harder items.

\paragraph{What to read for.}
Across cases we track five recurring patterns.
\textbf{(P1) Typed vs.\ free targets.} The policy never emits a natural-language query; it selects an action id whose type and target the environment validates.
\textbf{(P2) Same-state alternatives.} At each step the menu lists competitors that a free-query agent would rewrite into near-duplicate strings; the figures mark which competitors were rejected.
\textbf{(P3) Minimal committed sets.} $C_T$ retains only the sentences that entail the answer; near-name and period-mismatched distractors stay uncommitted even when they remain on the menu.
\textbf{(P4) Adaptive depth.} Bridge hops use \textsc{Lookup}; comparison and composition hops often terminate with zero lookups once operands are typed and visible.
\textbf{(P5) Credit locality.} On bridge hops, immediate answer likelihood may not rise until a later \textsc{Lookup}; SNC is the mechanism that still credits the enabling step via provenance.

\paragraph{How the cases map to claims.}
Bridge retrieval (Figure~\ref{fig:case-bridge}) is the canonical two-hop pattern where free-query aliasing and myopic credit both fail, so we reuse it for the free-query contrast and all three SNC panels.
Film-year, birth-date, and country comparison (Figures~\ref{fig:case-compare}, \ref{fig:case-birth}, \ref{fig:case-country}) stress early stop and distractor identity under different surface cues (near titles, shared first names, near-name opera titles).
MuSiQue composition (Figure~\ref{fig:case-compose}) stresses join-entity co-mention without relation-path walking.
Reading them as a set is intentional: no single figure shows every property, but the joint pattern matches the controlled ablations where swapping free query for the menu moves F1 by more than $17$ points and full SNC further improves multi-hop scores over leave-one-out variants (Tables~\ref{tab:interface}--\ref{tab:snc}).

\subsection{Menu Navigation: Process and Effectiveness}

\begin{figure*}[t]
\centering
\includegraphics[width=\textwidth]{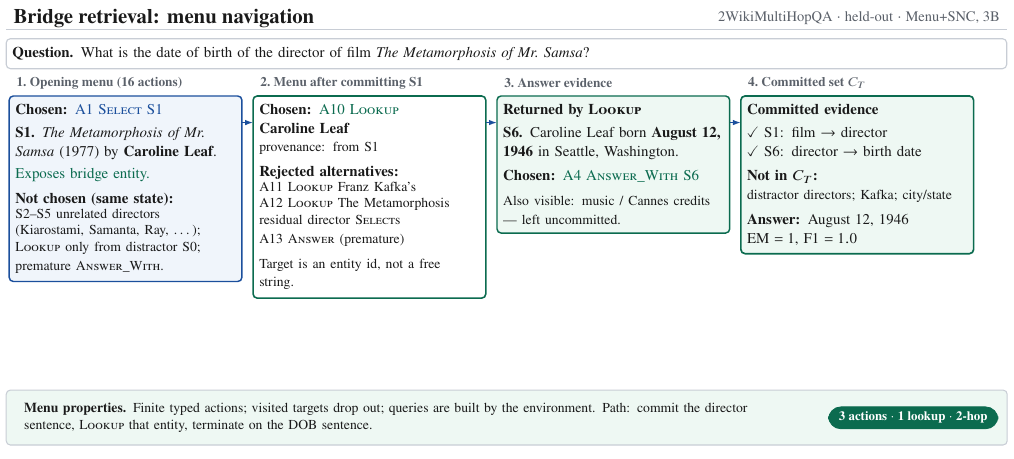}
\caption{2Wiki bridge retrieval: menu shows chosen vs.\ rejected targets at each hop.}
\label{fig:case-bridge}
\end{figure*}

\paragraph{Bridge retrieval.}
On the 2Wiki question in Figure~\ref{fig:case-bridge} (``What is the date of birth of the director of film \emph{The Metamorphosis Of Mr.\ Samsa}?''), the opening menu exposes $16$ typed actions, including four unrelated director biographies that share only loose lexical cues with the film title.
The policy selects \textbf{A1 \textsc{Select} S1} (film $\rightarrow$ Caroline Leaf), then from the post-commit menu issues \textbf{A10 \textsc{Lookup} Caroline Leaf} rather than \textsc{Lookup} on Franz Kafka or the film title, and terminates with \textbf{A4 \textsc{Answer\_With} S6} (birth date ``August 12, 1946'').
Both sentences remain in $C_T$; distractor directors never enter $C_T$.
The case exhibits the full menu pipeline in miniature.
First, initial dense retrieval surfaces a noisy candidate pool: lexical overlap with ``metamorphosis'' and literary names is enough to place Kafka-adjacent and other director bios on the menu, so the interface does not pretend that retrieval is already clean.
Second, \textsc{Select} commits the bridge sentence and expands the entity frontier: Caroline Leaf becomes a typed \textsc{Lookup} target rather than a free string the policy must re-type correctly.
Third, \textsc{Lookup} is an \emph{entity id}, so the environment builds the retrieval query $m_t=\mathrm{concat}(q,\mathrm{text}(C_t))$ and the policy cannot invent an alias string that collides with another director's retrieval class.
Fourth, \textsc{Answer\_With} closes when the attribute sentence is visible, so the final answer conditions only on committed evidence rather than on a free-form evidence bag that silently mixed distractors.
Relative to free query, the two hops are \emph{structurally distinct actions} rather than two paraphrases of ``who directed / when was she born,'' which is exactly the distinction that collapses under retrieval-equivalence classes in Figure~\ref{fig:diagnosis}.
The rejected alternatives also matter for training: each unused director bio, Kafka hop, or title hop becomes a same-state competitor for SNC, not a latent failure absent from the action log.

\begin{figure*}[t]
\centering
\includegraphics[width=\textwidth]{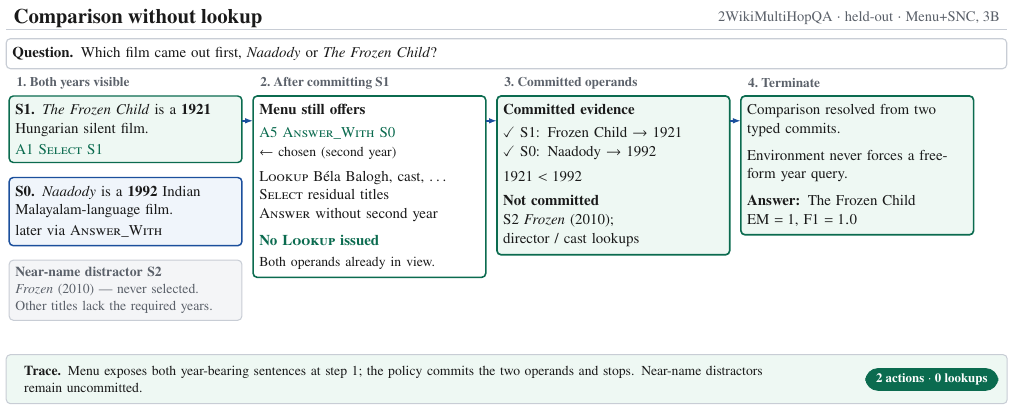}
\caption{2Wiki comparison: both operands committed without \textsc{Lookup}.}
\label{fig:case-compare}
\end{figure*}

\paragraph{Comparison without lookup.}
On the 2Wiki comparison in Figure~\ref{fig:case-compare}, both release years are already visible among the candidates.
The policy commits the two operands (\emph{The Frozen Child}, 1921; \emph{Naadody}, 1992), does not issue a \textsc{Lookup} on directors or cast, and terminates without committing the near-name distractor \emph{Frozen} (2010).
This case is important for \emph{interaction efficiency} (RQ4): the menu does not force a fixed number of retrieval hops.
When both operands are typed and visible, the optimal policy is two commits and stop; residual \textsc{Lookup}s on B\'{e}la Balogh or cast members remain offered but are correctly ignored.
That is a different competence from bridge retrieval: success here is not ``find more evidence,'' but ``recognize that the comparison is already closed and refuse extra search.''
The near-name distractor \emph{Frozen} (2010) illustrates verifiability: it is a different sentence id from either operand, so selecting it would be a different action with a different effect on $C_T$, not a silent contamination of a free-form evidence bag.
Under free query, a rewrite such as ``Frozen release year'' can land on the 2010 title by retrieval rank alone; under the menu, that contamination requires an explicit wrong \textsc{Select}, which the trained policy avoids.
The case also clarifies what the outcome reward alone sees: a correct year comparison with a polluted bag can still receive high F1 if the generated answer string is right, whereas the menu trajectory records that the distractor was never committed.

\begin{figure*}[t]
\centering
\includegraphics[width=\textwidth]{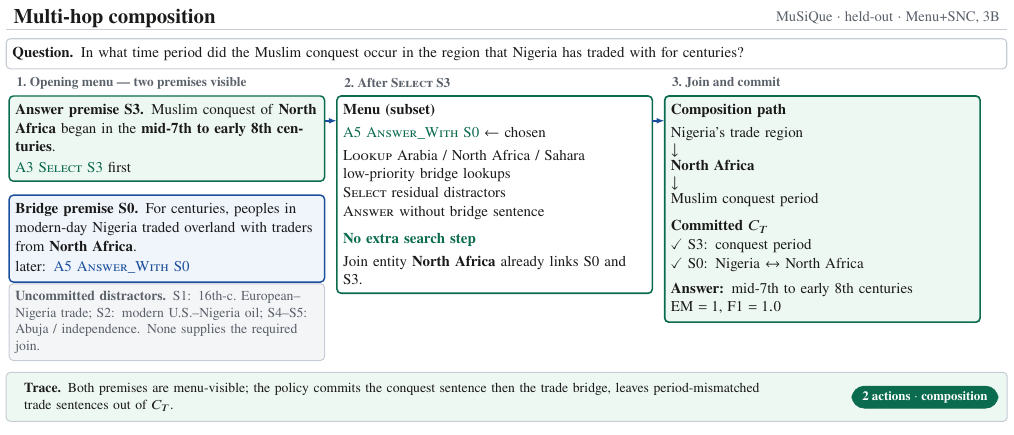}
\caption{MuSiQue composition: two committed premises, no extra search.}
\label{fig:case-compose}
\end{figure*}

\paragraph{Multi-hop composition.}
On the MuSiQue question in Figure~\ref{fig:case-compose}, the policy commits a bridge sentence (centuries-long overland trade between present-day Nigeria and North Africa) and an answer sentence (Muslim conquest of North Africa in the mid-7th to early 8th centuries), leaves period-mismatched trade sentences uncommitted, and answers without an additional search step.
After \textsc{Select} S3, the menu still offers residual \textsc{Lookup}s (Arabia, Sahara, low-priority bridges) and distractor \textsc{Select}s; the policy instead issues \textsc{Answer\_With} S0 to close the join on \emph{North Africa}.
Composition here is not path-walking over relation labels: the join entity is co-mentioned across two sentences, and the menu makes both premises selectable without requiring the policy to generate a second free-form query that re-states the join.
Uncommitted distractors (16th-century European trade; modern U.S.--Nigeria oil) show that the policy is not simply absorbing every trade-related sentence into $C_T$; sufficiency is judged relative to the required join rather than to lexical relatedness alone.
Two analytic consequences follow.
First, composition benefits from the same finiteness property as bridge retrieval, but the hard decision is \emph{which} visible premises to commit rather than \emph{which} entity to look up next.
Second, residual \textsc{Lookup}s after the first commit are temptation actions: they are legal, would change $U_{t+1}$, and would lengthen the trajectory, yet they are not needed once both premises of the join are visible.
Teaching the policy to ignore them is part of the efficiency story in RQ4 and cannot be read off from answer F1 alone.

\begin{figure*}[t]
\centering
\includegraphics[width=\textwidth]{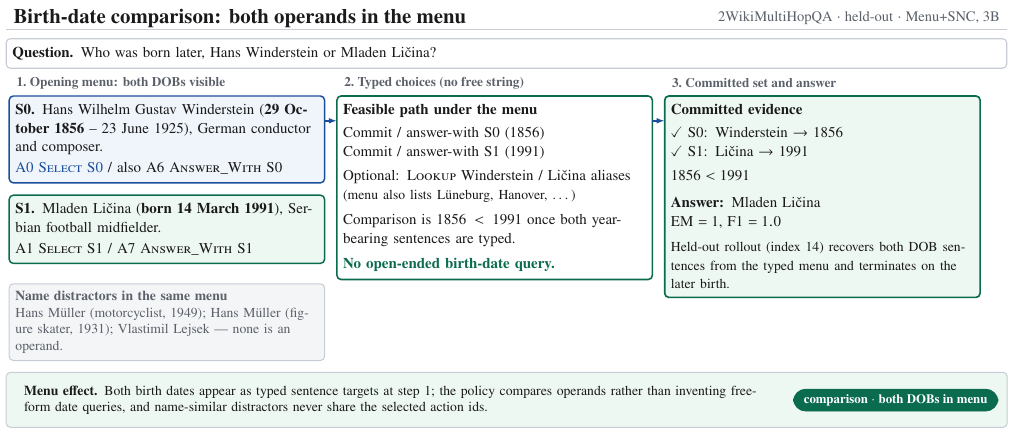}
\caption{2Wiki birth-date comparison: both DOBs exposed as typed menu targets.}
\label{fig:case-birth}
\end{figure*}

\paragraph{Birth-date comparison.}
On the 2Wiki question in Figure~\ref{fig:case-birth}, both birth dates are already typed menu targets at step~1 (Winderstein, 1856; Li\v{c}ina, 1991), alongside name-similar distractors (e.g., Hans M\"{u}ller motorcyclist / figure skater).
The policy compares the two operands and answers \emph{Mladen Li\v{c}ina} without inventing free-form date queries.
This case stresses a different failure mode than film-year comparison: distractors share the first name \emph{Hans} and are otherwise plausible biography sentences, so lexical retrieval is ambiguous even when the gold names are fully specified in the question.
Under free query, ``when was Hans \ldots born'' is an underspecified string that can retrieve any of them; under the menu, each biography is a distinct \textsc{Select}/\textsc{Answer\_With} target, so the policy's choice is an explicit identity decision rather than a hope that dense retrieval ranked the right person first.
The comparison also makes the answer predicate operational without an extra hop: both DOBs are already visible, so the policy's work is relational (who is younger) rather than informational (fetch a missing date).
That separation matters for credit assignment: myopic gains on ``fetch something'' would not explain the decision, whereas the menu forces the decision to be about which typed operands enter $C_T$.

\begin{figure*}[t]
\centering
\includegraphics[width=\textwidth]{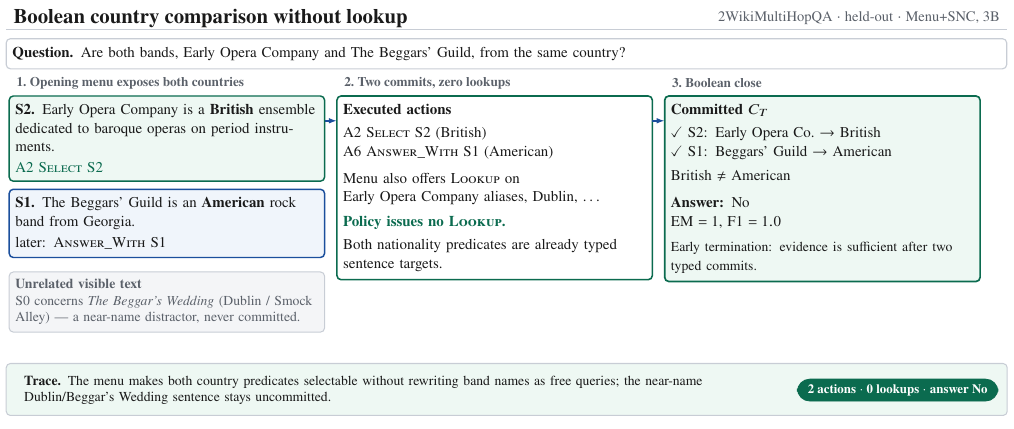}
\caption{2Wiki country comparison: two nationality commits, zero lookups.}
\label{fig:case-country}
\end{figure*}

\paragraph{Boolean country comparison.}
On the 2Wiki question in Figure~\ref{fig:case-country}, the menu exposes a British Early Opera Company sentence and an American Beggars' Guild sentence.
The policy commits both, issues no \textsc{Lookup}, leaves the near-name Dublin/\emph{Beggar's Wedding} distractor uncommitted, and answers \emph{No}.
Boolean multi-hop questions are easy to get ``half right'': retrieving one band's country and guessing the other, or matching only the more distinctive name.
The menu makes both nationality predicates first-class, and early termination again appears---two commits suffice.
The near-name distractor (ballad opera / Dublin theatre history) would be a typical free-query attractor for ``Beggar,'' yet it never shares the selected action ids of the rock band sentence.
Analytically, this case combines the strengths of the film-year and birth-date comparisons: operands are fully visible (no lookup), distractors are identity confusions rather than missing attributes, and the answer is a closed boolean rather than a copied span.
A polluted free-form bag containing the Dublin theatre sentence could flip one nationality predicate without an explicit wrong action; the menu instead leaves this failure mode as an unused competitor.

\paragraph{Cross-case menu analysis.}
The five menu trajectories jointly instantiate the three constructive properties of the menu environment.
\emph{Finiteness:} every decision is over a bounded set of action ids (here, on the order of $10$--$16$ at the opening step), not $\Sigma^*$.
\emph{Verifiability:} rejected distractors (unrelated directors, \emph{Frozen} 2010, Hans M\"{u}ller, \emph{Beggar's Wedding}) are explicit competing actions rather than latent retrieval noise.
\emph{Previewability:} although the inference-time policy does not call the scorer, every rejected alternative is a deterministic index operation that SNC could have expanded read-only---the same structure used at training time.
A second cross-cutting observation is \emph{adaptive depth}: bridge retrieval uses one \textsc{Lookup}; the four comparison / composition cases use zero.
That pattern is consistent with the interaction-efficiency results (RQ4): the interface permits short trajectories when evidence is already typed, rather than always spending a fixed search budget.
Third, committed sets are minimal. Across cases $C_T$ contains exactly the sentences needed for entailment, which is the form of evidence the outcome F1 reward alone cannot teach without intermediate structure: a correct answer string can still be produced from an over-committed bag, yet the logged trajectory records which premises were actually committed and when the policy stopped.
Fourth, the hard decisions differ by behavior class.
Bridge retrieval's hard step is choosing the right entity to look up after the bridge commit; comparison's hard step is stopping and rejecting near-name operands; composition's hard step is committing the right join premises while ignoring period-mismatched trade sentences.
A free-query agent collapses all three into ``emit another string and hope retrieval ranks well''; the menu forces each class to surface as a different typed choice.
Thus, Table~\ref{tab:interface} shows large menu gains even with outcome-only credit: the interface changes the decision problem before process rewards.

\subsection{Interface Contrast: Free Query vs.\ Menu}

\begin{figure*}[t]
\centering
\includegraphics[width=\textwidth]{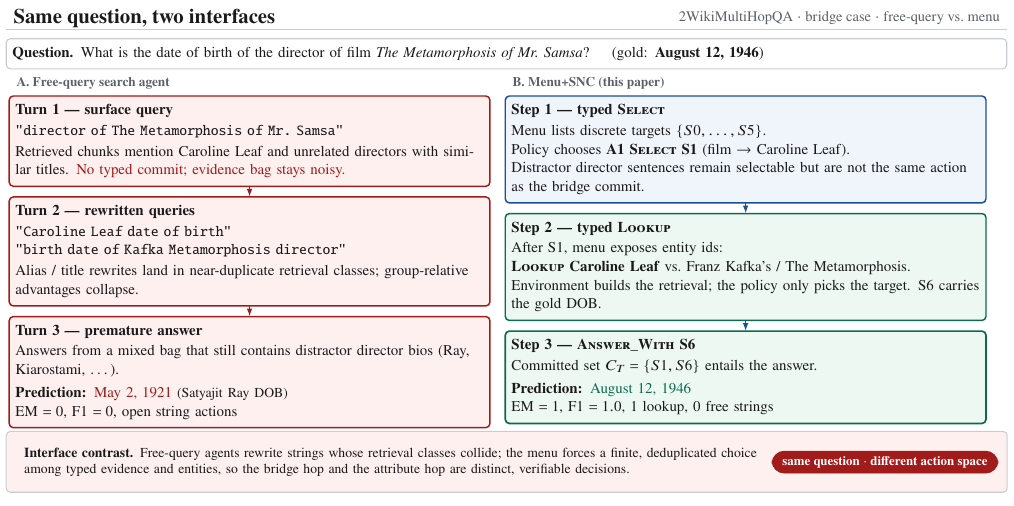}
\caption{Same 2Wiki bridge question: free-query failure vs.\ menu success.}
\label{fig:case-freequery}
\end{figure*}

\paragraph{Free-query contrast on the bridge question.}
Figure~\ref{fig:case-freequery} revisits the Caroline Leaf bridge question under a free-form query interface.
Surface rewrites (``director of \ldots'', ``Caroline Leaf date of birth'', title paraphrases) land in near-duplicate retrieval classes that still mix distractor director biographies; the agent answers with an incorrect birth date drawn from a distractor bio (Satyajit Ray).
Under the menu, the same corpus evidence is reparameterized as typed \textsc{Select}/\textsc{Lookup}/\textsc{Answer\_With} decisions with explicit targets, so the bridge hop and the attribute hop remain distinct, verifiable actions and the gold DOB is recovered in three steps.
The qualitative failure matches the quantitative diagnosis in Figure~\ref{fig:diagnosis}: free-query groups lose retrieval-distinct diversity even when surface strings look diverse, and group-relative advantages then compare near-clones.
Two interface mechanisms explain the gap.
First, free-query aliases are many-to-one into retrieval classes: several phrasings of the director question can retrieve the same mixed candidate page, so GRPO's within-group baseline has little genuine contrast.
Second, free-query evidence bags are not action-indexed: a distractor birth date can enter the conditioning context without any logged decision that says ``commit Ray's bio rather than Leaf's.''
The menu does not magically remove distractor text from the corpus; it prevents distractors from being \emph{the same action} as the bridge commit and from entering $C_T$ without an explicit \textsc{Select}/\textsc{Answer\_With}.
That is the interface-level claim of RQ2: replacing free query with the menu accounts for the large F1 gaps in Table~\ref{tab:interface} (more than $17$ F1 points overall, and more than $35$ on MuSiQue under outcome-only training), beyond what denser free-query credit (IGPO) can recover.
IGPO can densify process signal inside free query, but it cannot enumerate a same-state frontier of typed alternatives or block aliasing at the action level; the residual free-query gap after IGPO is therefore expected, not an optimization failure.

\subsection{SNC: Frontier, Enablement, and Credit Comparison}

\begin{figure*}[t]
\centering
\includegraphics[width=\textwidth]{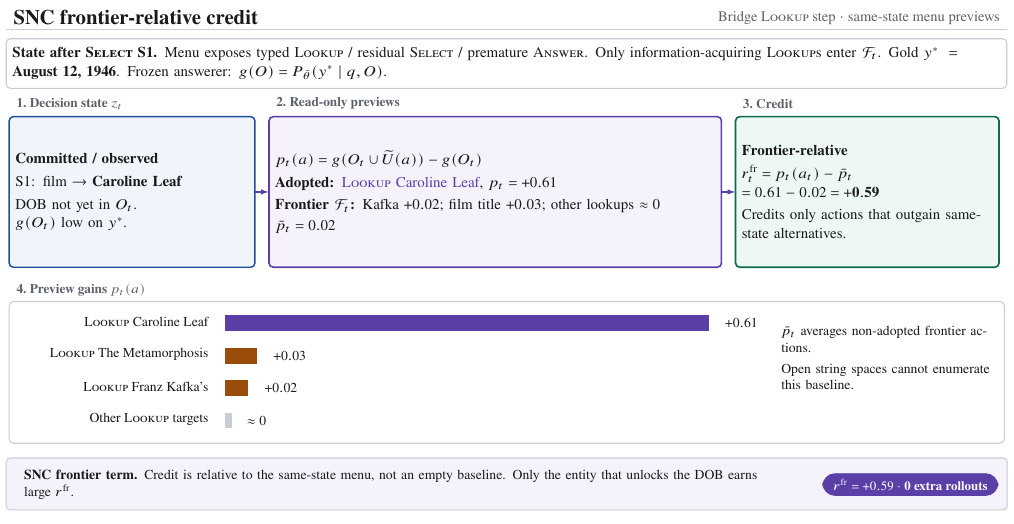}
\caption{SNC frontier term on the bridge \textsc{Lookup}: gain bars vs.\ same-state alternatives.}
\label{fig:case-snc-fr}
\end{figure*}

\paragraph{SNC frontier-relative credit.}
Figure~\ref{fig:case-snc-fr} scores the post-\textsc{Select} menu of the bridge trajectory with the frozen answerer.
\textsc{Lookup Caroline Leaf} yields a large preview gain on the gold birth date ($p_t{=}{+}0.61$), while the same-state frontier---\textsc{Lookup} Kafka / film title and other \textsc{Lookup} targets---yields near-zero gain; residual \textsc{Select}s and premature \textsc{Answer} remain menu actions but are excluded from $\mathcal{F}_t$.
The frontier baseline $\bar p_t$ is the mean of those alternatives, so $r_t^{\mathrm{fr}}{=}p_t(a_t)-\bar p_t{=}{+}0.59$ credits only the action that outgains genuine competitors at the same state.
Several analytic points follow.
First, the baseline is \emph{in-menu}, not empty-context: the policy is not rewarded merely for retrieving something, but for beating alternatives that were actually feasible at that step.
Second, superficial entity hops (Kafka, film title) receive near-zero $p_t$ even though they are valid menu entries---so $r^{\mathrm{fr}}$ is not a generic ``use \textsc{Lookup}'' bonus, nor a length bonus for taking an extra hop.
Third, computing this term requires only read-only previews and a frozen scorer on the realized trajectory; it needs no tree search and no extra agent rollouts, which is the cost claim in Table~\ref{tab:cost}.
Fourth, the same menu structure that makes inference decisions typed also makes the training baseline well-defined: open string spaces cannot enumerate $\mathcal{F}_t$ at all, so the menu is a precondition for frontier-relative credit, not an optional extra.
In Table~\ref{tab:snc}, removing $r^{\mathrm{fr}}$ while keeping enablement still hurts multi-hop F1: without a same-state baseline, the gold \textsc{Lookup}'s large $p_t$ is harder to separate from weaker positive signals.

\begin{figure*}[t]
\centering
\includegraphics[width=\textwidth]{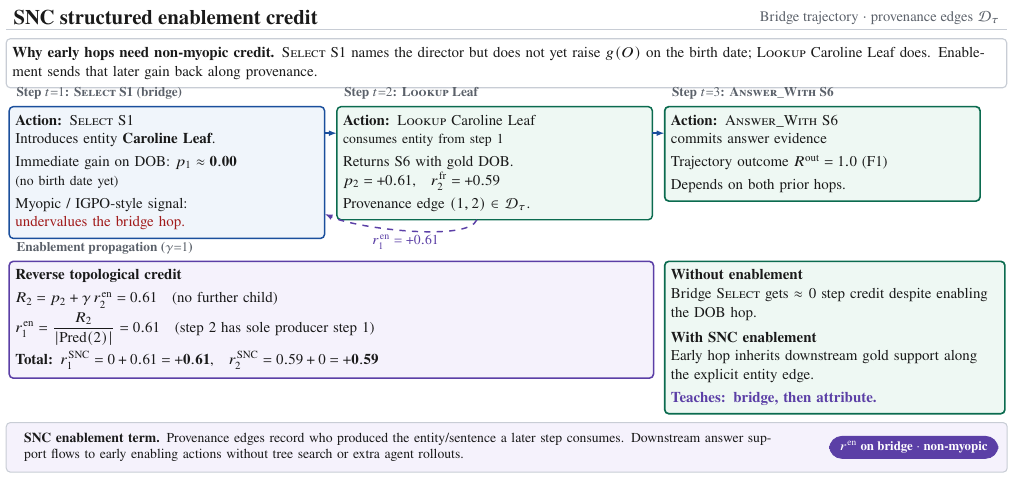}
\caption{SNC enablement term: downstream DOB gain flows back to the bridge \textsc{Select}.}
\label{fig:case-snc-en}
\end{figure*}

\paragraph{SNC enablement credit.}
Figure~\ref{fig:case-snc-en} shows why the early bridge hop needs non-myopic credit.
\textsc{Select} S1 exposes Caroline Leaf but does not yet raise $g(O)$ on the birth date ($p_1{\approx}0$); the subsequent \textsc{Lookup} consumes that entity and realizes $p_2{=}{+}0.61$.
SNC records the provenance edge $(1,2)\in\mathcal{D}_\tau$ and propagates $r_1^{\mathrm{en}}{=}R_2/|\mathrm{Pred}(2)|{=}0.61$ to step~1, yielding $r_1^{\mathrm{SNC}}{=}{+}0.61$ and $r_2^{\mathrm{SNC}}{=}{+}0.59$.
Without enablement, the necessary bridge hop would receive near-zero process credit despite enabling the gold attribute retrieval.
This is the multi-hop credit pathology that pure outcome rewards and per-step marginal gains both miss: the first hop is \emph{causally} necessary but \emph{locally} uninformative for $y^*$.
Provenance edges are environment bookkeeping (which earlier action produced the entity a later step consumed), so the back-flow does not require a learned value model or counterfactual rollouts from intermediate states.
Leave-one-out ablation supports the same reading: removing enablement while keeping the frontier term drops multi-hop F1 (Table~\ref{tab:snc}), and the drop is largest where delayed bridge structure is common.
The panel makes the mechanism explicit: myopic credit nearly misses step~1 even though it enables step~2.

\begin{figure*}[t]
\centering
\includegraphics[width=\textwidth]{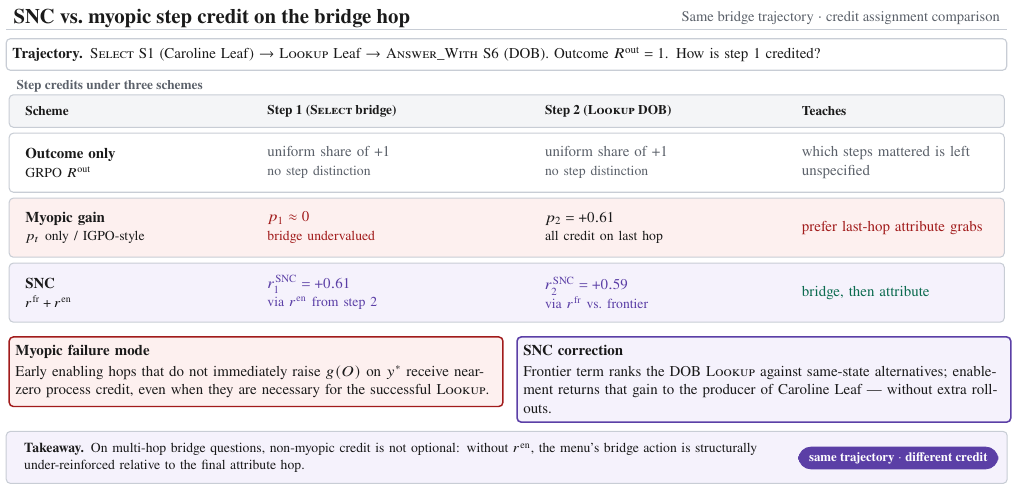}
\caption{Same bridge trajectory under outcome-only, myopic, and SNC credit.}
\label{fig:case-snc-myopic}
\end{figure*}

\paragraph{SNC vs.\ myopic credit.}
Figure~\ref{fig:case-snc-myopic} places three credit schemes on the same bridge trajectory.
Outcome-only GRPO spreads $R^{\mathrm{out}}$ uniformly across tokens and does not distinguish which hop mattered; both the bridge commit and the attribute lookup receive the same trajectory-level success signal, diluted across all tokens.
Myopic / IGPO-style gains assign $p_1{\approx}0$ to the bridge \textsc{Select} and concentrate process credit on the final \textsc{Lookup}, teaching last-hop attribute grabs and under-rewarding the enabling hop that created the entity frontier.
SNC restores $r_1^{\mathrm{SNC}}{=}{+}0.61$ via enablement while keeping a large frontier-relative term on the DOB hop, teaching the non-myopic pattern ``bridge, then attribute.''
The figure is the qualitative counterpart of the SNC leave-one-out ablation (Table~\ref{tab:snc}): removing the frontier term loses discrimination among same-state alternatives; removing enablement loses credit for early hops; $\lambda{=}0$ collapses to outcome-only and drops F1 by $3.08$ / $4.55$ / $2.88$ on 2Wiki / HotpotQA / MuSiQue.
Because both SNC terms are computed only when the menu makes alternatives and provenance explicit, the ablation gap is an interface$\times$credit interaction, not a free-standing reward trick that could be ported unchanged to free query.
Reading the three schemes on one trajectory also clarifies a common confusion: SNC is not ``more reward.''
On this path the total process mass is structured, not merely larger---credit is moved onto the enabling hop and concentrated on the uniquely good same-state alternative, which is what the leave-one-out variants disentangle.

\subsection{Additional Failure Cases}
\label{app:cases-fail}

We also report three held-out HotpotQA trajectories of the same Menu+SNC Qwen2.5-3B checkpoint used in Table~\ref{tab:main}, on which the final answer is incorrect.
As in the success cases, every step is a typed menu action under the full Harness-G configuration, so the trajectories remain fully inspectable even when the prediction misses the gold string.

\begin{figure*}[t]
\centering
\includegraphics[width=\textwidth]{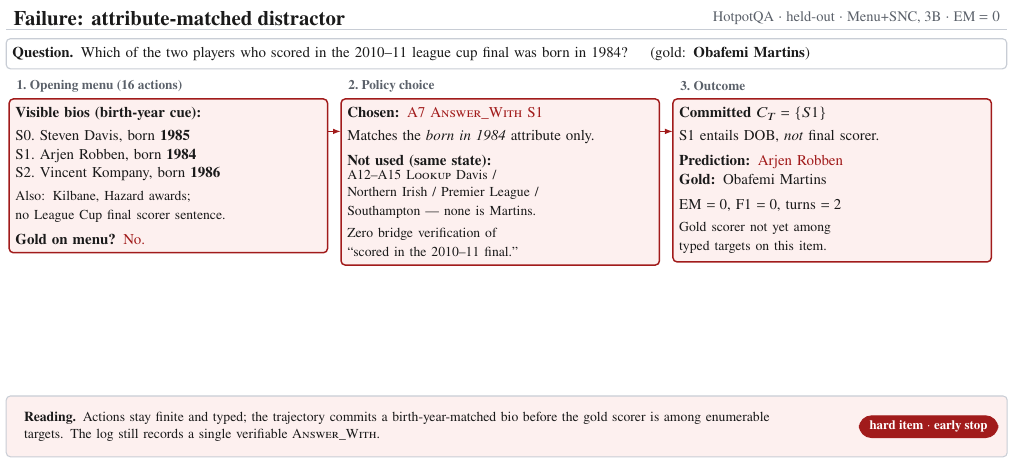}
\caption{HotpotQA failure case: \textsc{Answer\_With} on a birth-year-matched candidate; gold scorer is not yet among menu targets.}
\label{fig:case-fail-attr}
\end{figure*}

\paragraph{Attribute-matched distractor.}
On the HotpotQA question in Figure~\ref{fig:case-fail-attr} (``Which of the two players who scored in the 2010--11 league cup final was born in 1984?''), the opening menu lists footballer biographies whose birth years straddle 1984 (Davis 1985, Robben 1984, Kompany 1986) but does not yet surface Obafemi Martins or a League Cup final sentence.
The policy issues \textbf{A7 \textsc{Answer\_With} S1} (Arjen Robben, born 1984) at turn~1, following the birth-year cue that is explicitly typed on the menu.
Available \textsc{Lookup} targets at that state (Davis, Northern Irish, Premier League, Southampton) likewise do not include the gold entity.
EM $=0$, F1 $=0$, two turns.
The typed log still records a single, verifiable commit; the missed hop is that the gold scorer was not yet among the enumerable targets for this item.

\begin{figure*}[t]
\centering
\includegraphics[width=\textwidth]{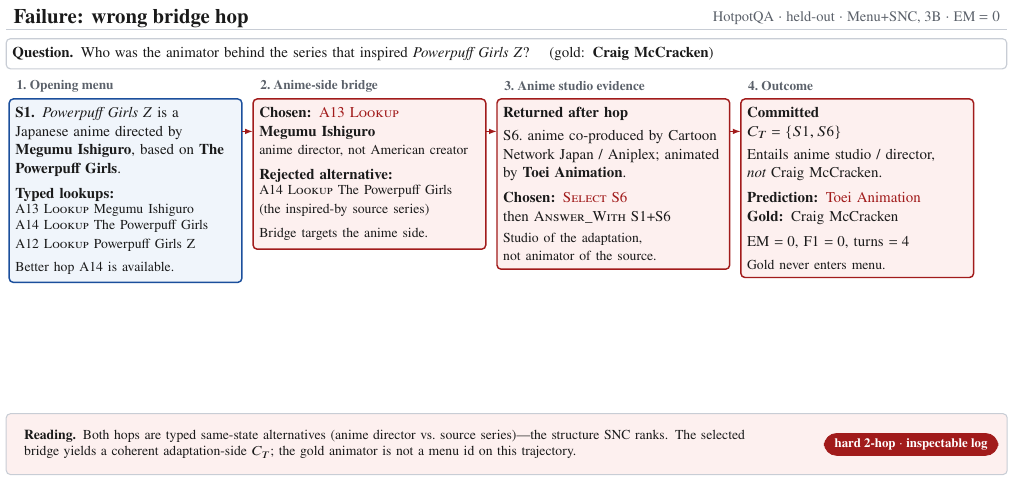}
\caption{HotpotQA failure case: anime-side bridge \textsc{Lookup} among two typed alternatives; gold animator is not on the menu of this trajectory.}
\label{fig:case-fail-bridge}
\end{figure*}

\paragraph{Competing bridge lookups.}
On the question in Figure~\ref{fig:case-fail-bridge} (``Who was the animator behind the series that inspired \emph{Powerpuff Girls Z}?''), opening sentence S1 already states that the anime is based on \emph{The Powerpuff Girls} and exposes two competing typed lookups: \textbf{A13 \textsc{Lookup} Megumu Ishiguro} (anime director) vs.\ \textbf{A14 \textsc{Lookup} The Powerpuff Girls} (source series).
The policy chooses the anime-side entity, commits the Toei Animation production sentence, and answers ``Toei Animation'' (EM $=0$, F1 $=0$, four turns).
The gold animator Craig McCracken is not among the visible sentences or entity ids on this trajectory.
Both hops remain loggable same-state alternatives of the kind SNC is designed to rank; the case simply shows a hard two-hop item where the selected bridge leads to a coherent adaptation-side $C_T$ rather than the source-series answer.

\begin{figure*}[t]
\centering
\includegraphics[width=\textwidth]{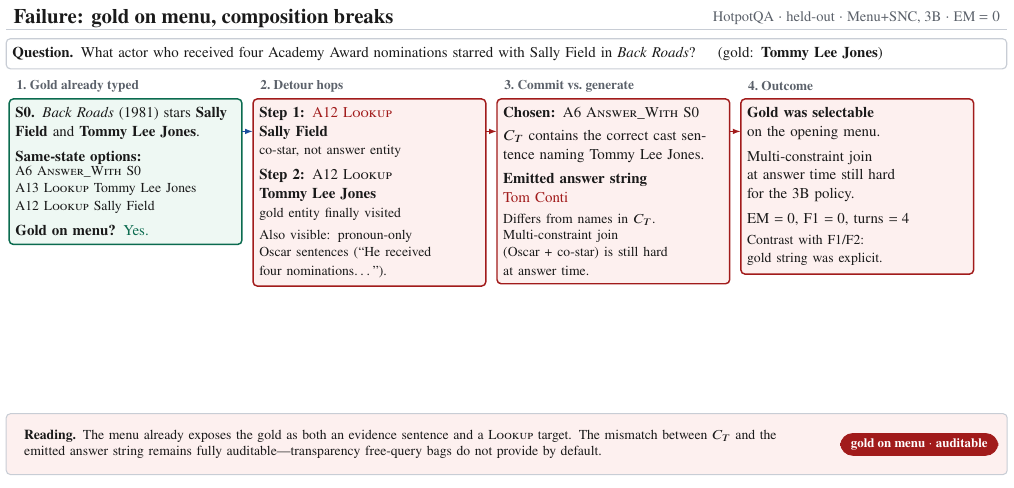}
\caption{HotpotQA failure case: gold name and \textsc{Answer\_With} are available on the opening menu; the emitted answer string still mismatches $C_T$.}
\label{fig:case-fail-coref}
\end{figure*}

\paragraph{Gold visible, answer string still mismatches.}
Figure~\ref{fig:case-fail-coref} shows a complementary pattern in which the gold name is already on the menu.
For ``What actor who received four Academy Award nominations starred with Sally Field in \emph{Back Roads}?'', opening S0 explicitly names \textbf{Tommy Lee Jones}, and the menu offers both \textbf{A6 \textsc{Answer\_With} S0} and \textbf{A13 \textsc{Lookup} Tommy Lee Jones}.
The policy looks up Sally Field, later looks up Tommy Lee Jones, commits S0, and generates \textbf{Tom Conti} (EM $=0$, F1 $=0$).
Pronoun-only Oscar sentences remain visible alongside the cast sentence.
Here the menu still makes the decision sequence and committed evidence fully auditable: the correct cast sentence was selectable, and the mismatch between $C_T$ and the emitted answer string is explicit in the log---a form of transparency that free-query bags do not provide by default.

\paragraph{Reading the three cases together.}
Across Figures~\ref{fig:case-fail-attr}--\ref{fig:case-fail-coref}, two patterns recur.
In the first two, the gold entity is not yet among typed targets when the trajectory terminates; in the third, the gold string is selectable and the residual difficulty is multi-constraint joining at answer time.
In all three, actions remain finite, non-aliased, and recorded---the same interface properties illustrated by the success cases---so incorrect outcomes stay localizable to candidate coverage or answer-string composition rather than to an opaque free-query rewrite.
These examples are therefore best read as complementary qualitative traces: they show where denser initial candidates, stronger multi-constraint answering, or larger backbones could further improve end-task F1 while preserving the menu and SNC machinery.

\subsection{Synthesis}
\label{app:cases-synth}

Reading the success cases left-to-right, the menu answers \emph{what} the policy may do (finite, typed, distractor-separated decisions; adaptive depth; minimal $C_T$), while SNC answers \emph{how} those decisions are reinforced when payoffs are delayed (frontier ranking at the attribute hop; enablement back to the bridge hop).
Free-query contrast shows that densifying credit inside an open string space does not recreate either property: the same question still collapses into aliased retrieval and a distractor DOB, consistent with the residual free-query gap after IGPO in Table~\ref{tab:interface}.
These qualitative mechanisms align with the controlled RQ2 interface and SNC ablations: large gains from replacing free query by the menu, and additional multi-hop gains from full SNC over outcome-only or leave-one-out variants.
The additional failure cases in \S\ref{app:cases-fail} keep the same typed action log on harder held-out items and show that incorrect outcomes remain inspectable at the level of individual menu decisions.
The cases also clarify a negative claim: Harness-G is not ``a better dense retriever.''
In every success figure the corpus still contains distractors; initial menus still surface noisy candidates; the contribution is that distractors become \emph{rejected actions} and delayed payoffs become \emph{explicit step credits}, both of which free-query RL leaves ill-posed.
A second negative claim follows for reward design alone: process densification without a typed frontier (IGPO under free query) and outcome-only training under the menu each leave a residual failure mode that the complementary component repairs---aliasing for the former, delayed bridge credit for the latter.
The appendix therefore supports the paper's joint thesis: interface reparameterization and structured non-myopic credit are coupled, and the held-out trajectories---both successful and unsuccessful---show that coupling at the level of individual decisions rather than only at aggregate F1.

\section{Environment Protocol and Prompts}
\label{app:prompts}

\begin{figure*}[t]
\centering
\includegraphics[width=\textwidth]{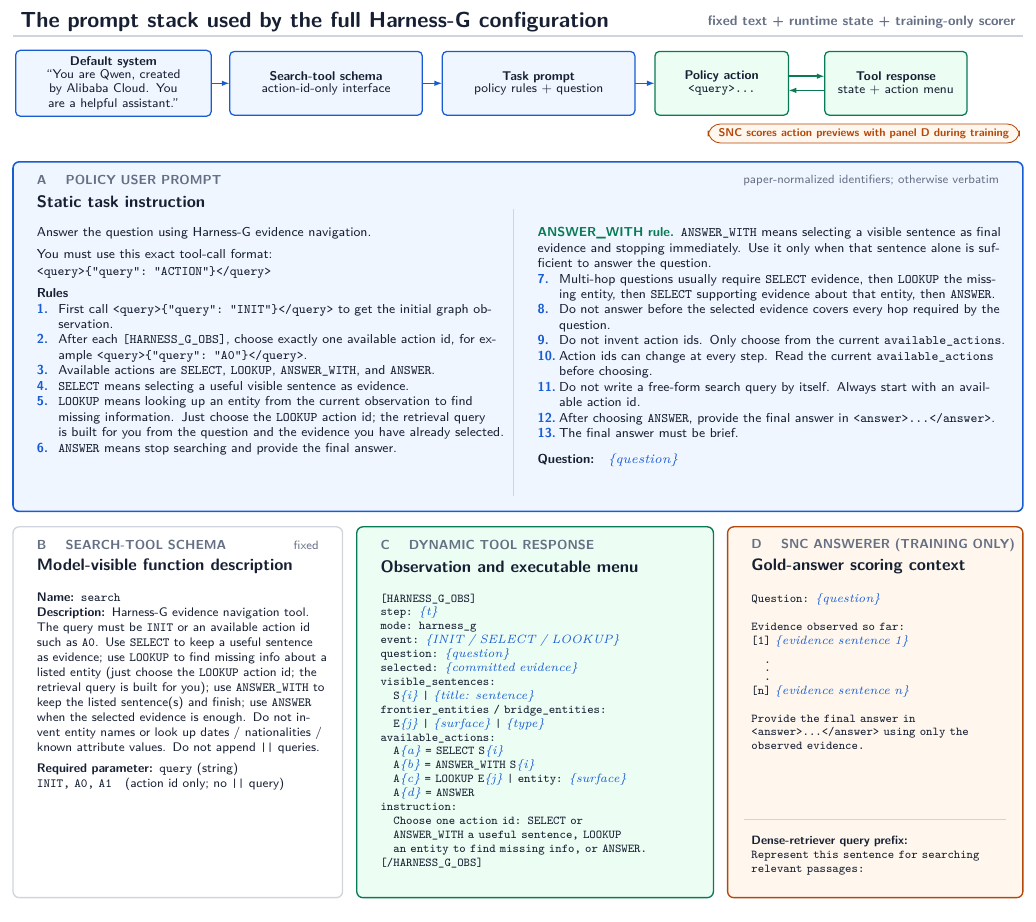}
\caption{Prompt stack for the full Harness-G configuration.}
\label{fig:prompt-stack}
\end{figure*}

The full Harness-G configuration uses the same prompt stack on all six benchmarks.
Figure~\ref{fig:prompt-stack} consolidates every task-specific text component: the policy instruction, the model-visible search-tool schema, the event-conditioned environment response, and the frozen-answerer prompt used by SNC.
We retain the default Qwen2.5 system message rather than adding a custom system prompt.
The SNC scorer prompt only computes teacher-forced gold-answer likelihoods during training and never enters the policy context.

\paragraph{Observation format.}
At each step, the environment renders a delimited text observation with:
\begin{itemize}
    \item the step index and original question;
    \item the numbered committed-evidence list;
    \item the currently visible sentences in the form \texttt{S0 | title: sentence};
    \item bridge and frontier entities exposed by the current state; and
    \item the executable action menu with identifiers such as \texttt{A0 = SELECT S0}, \texttt{A1 = LOOKUP E3 | entity: ...}, \texttt{A2 = ANSWER\_WITH S0}, and \texttt{A3 = ANSWER}.
\end{itemize}
The policy must choose exactly one available action id and may not issue a natural-language search query.

\paragraph{Deterministic query construction and validation.}
The policy never supplies retrieval text.
For \textsc{Lookup}, the environment concatenates the question and committed evidence into a query capped at 64 words, gathers candidates from entity mentions, synonym links, and sentence adjacency, and ranks them with the dense encoder.
Any free-text rider appended to an action, including a \texttt{||}-suffixed query, is rejected.
Committed sentences are removed from future \textsc{Select} choices, visited entities are removed from future \textsc{Lookup} choices, and duplicate retrievals are merged before the next menu is shown.
The tool description also instructs the policy not to invent entities or request dates, nationalities, and already-known attribute values.

% Table~\ref{tab:build} appears in the main paper (RQ4).

% Open-corpus appendix material remains deferred; the zero-advantage
% comparison is reported in the main text (Figure~\ref{fig:zadv}).

\end{document}